\pdfoutput=1

\documentclass[11pt]{article}

\usepackage[]{acl}

\usepackage{times}
\usepackage{latexsym}
\usepackage{url}
\usepackage[T1]{fontenc}
\usepackage{microtype}
\usepackage{multirow}
\usepackage{caption}
\usepackage{subcaption}
\usepackage{booktabs}
\usepackage{graphicx}
\usepackage{amsmath}
\usepackage{empheq}
\usepackage{times}
\usepackage{latexsym}
\usepackage{amssymb}
\usepackage{colortbl}
\usepackage{array}

\usepackage[T1]{fontenc}

\usepackage[utf8]{inputenc}

\usepackage{microtype}

\title{A Comprehensive Analysis of Adapter Efficiency}

\author{
    Nandini Mundra$^{1,2}$\thanks{~Corresponding author: Nandini Mundra (\href{mailto:cs21s041@cse.iitm.ac.in}{cs21s041@cse.iitm.ac.in})} \hspace{0.2cm} Sumanth Doddapaneni$^{1,2}$ \hspace{0.2cm} Raj Dabre$^{3}$ \hspace{0.2cm} \\
    \textbf{Anoop Kunchukuttan}$^{1,2,4}$ \hspace{0.2cm} \textbf{Ratish Puduppully}$^{5}$ \hspace{0.2cm} \textbf{Mitesh M. Khapra}$^{1,2}$ 
    \\ \\
    $^1$Indian Institute of Technology, Madras \hspace{0.2cm} 
    $^2$AI4Bharat \\ 
    $^3$ National Institute of Information and Communications Technology \hspace{0.2cm}
    $^4$Microsoft \hspace{0.2cm} \\
    $^5$Institute for Infocomm Research (I$^2$R), A$^{*}$STAR, Singapore \hspace{0.2cm} \\
}

\begin{document}
\maketitle
\begin{abstract}
Adapters have been positioned as a parameter-efficient fine-tuning (PEFT) approach, whereby a minimal number of parameters are added to the model and fine-tuned. However, adapters have not been sufficiently analyzed to understand if PEFT translates to benefits in training/deployment efficiency and maintainability/extensibility. Through extensive experiments on many adapters, tasks, and languages in supervised and cross-lingual zero-shot settings, we clearly show that for Natural Language Understanding (NLU) tasks, the parameter efficiency in adapters does not translate to efficiency gains compared to full fine-tuning of models. More precisely, adapters are relatively expensive to train and have slightly higher deployment latency. Furthermore, the maintainability/extensibility benefits of adapters can be achieved with simpler approaches like multi-task training via full fine-tuning, which also provide relatively faster training times. We, therefore, recommend that for moderately sized models for NLU tasks, practitioners should rely on full fine-tuning or multi-task training rather than using adapters. Our code is available at \url{https://github.com/AI4Bharat/adapter-efficiency}.
\end{abstract}

\section{Introduction}

Pretraining followed by fine-tuning \cite{devlin-etal-2019-bert,DBLP:journals/corr/abs-1907-11692} is the most commonly used paradigm in NLP, but as pre-trained models grow in size, fine-tuning the entire model (full fine-tuning) becomes costly. Maintaining a copy of the model for each task is costly, and parameter efficient fine-tuning (PEFT) has become an active area of research that focuses on fine-tuning a minimal number of parameters while still achieving comparable performance as of full fine-tuning. Fine-tuning adapters \cite{pmlr-v97-houlsby19a}, which typically involves fine-tuning tiny feed-forward layers injected into the model, is the most popular PEFT approach. Given the significantly lesser number of parameters that need to be fine-tuned, adapters are very useful in situations where the pre-trained model is too large to perform fine-tuning of all its parameters. Furthermore, the availability of frameworks such as Adapter-hub \cite{pfeiffer-etal-2020-adapterhub}, which is built on top of Transformers \cite{wolf-etal-2020-transformers}, has made it easy for researchers to experiment with PEFT methods and deploy their models.

\begin{figure}[t!]
    \centering
    \includegraphics[width=\columnwidth]{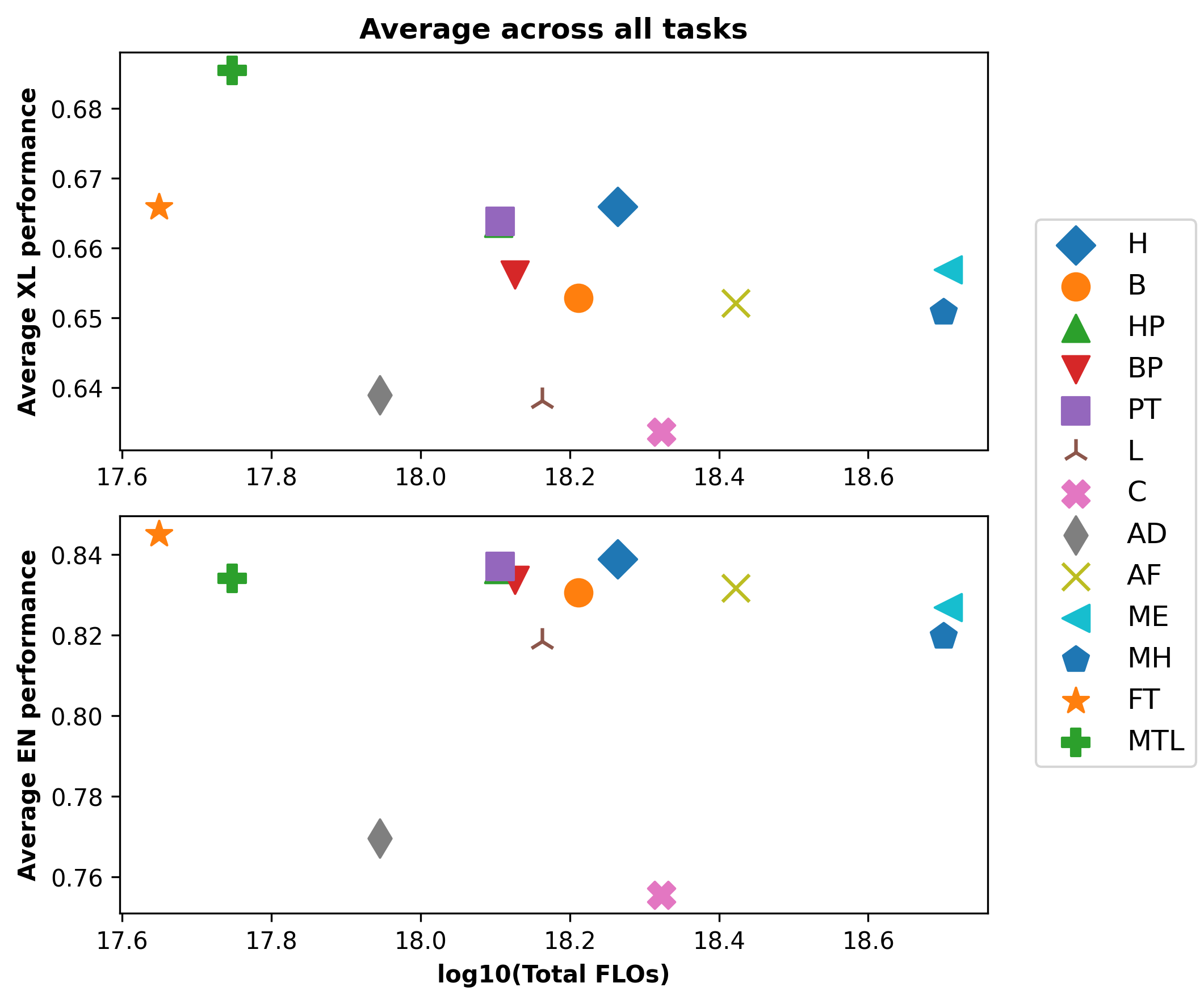}
    \caption{A comparison of 10 different adapters with simpler baselines like full fine-tuning (FT) and multi-task learning (MTL). In the top figure the y-axis shows the zero-shot performance averaged across all tasks and all languages. In the bottom figure, the y-axis shows the En performance averaged across all tasks. The abbreviations used are-`\textbf{H}' - Houlsby, `\textbf{B}' - Bapna, `\textbf{HP}' - Houlsby Parallel\footnotemark, `\textbf{BP}'- Bapna Parallel, `\textbf{PT}'- Prefix Tuning, `\textbf{L}'- LoRA, `\textbf{C}' - Compacter, `\textbf{AD}'- Adapter Drop, `\textbf{AF}' - Adapter Fusion, `\textbf{ME}' - MADX-en, `\textbf{MH}' - MADX-hi, `\textbf{FT}' - Fine-tuning, `\textbf{MTL}'- Multi-task-learning. }    
    \label{fig:cc-dist}
\end{figure}

\footnotetext{HP is overlapped by PT in this figure.}

While adapters are clearly parameter efficient, we argue that, in practice, there is more to efficiency than just the number of parameters being fine-tuned. For example, a parameter-efficient model will require more floating-point operations (FLOs), owing to the additional parameters added and this will affect latency. Additionally, the number of steps till convergence will lead to compute inefficiency - we find that adapters take more steps to converge as compared to full fine-tuning. 
Although adapters can be easily used to extend an existing model to new tasks, efficiency in terms of the total cost over incorporating multiple tasks is often not studied.
We thus believe that a thorough study of adapters in comparison with simpler baselines is needed to answer the following question: \textbf{\textit{What are adapters really efficient at?}} 

We recommend that to answer this question one should look beyond the number of parameters and consider other indicators of efficiency, such as, (i) training time and compute (FLOs), (ii) deployability via inference latency (iii)  and maintainability. Existing studies have looked at one or more of the above metrics but a thorough study comparing multiple popular adapters on different tasks across languages, especially in a cross-lingual setting, is missing. 
A simpler baseline is multi-task learning (MTL) \cite{liu-etal-2019-multi}, where a single model is jointly trained for all tasks via task specific classification heads.
Most works on adapters do not compare against MTL, making it hard to get a clear picture of the real utility of adapters.

In this work, we try to build a clearer picture by experimenting with 10 different adapters and 6 Natural Language Understanding (NLU) tasks spanning 11 Indian languages. We focus on zero-shot transfer, wherein we fine-tune models only on the English training data. We compare adapters with full fine-tuning and multi-task learning (MTL) and find that, quite contrary to popular beliefs, these simpler baselines are more efficient along multiple axes. Our work also lays down a framework for evaluating adapters along multiple dimensions. The key findings of our work along these dimensions, as summarized in Figure~\ref{fig:cc-dist} are as follows: 

\noindent \textbf{Compute efficiency:} Adapters are compute inefficient and need on average 325.6\% more compute (measured in FLOs) than full fine-tuning, mainly because they take 20.2\% longer time to converge. 

\noindent \textbf{Inference overhead:} Adapters insert new layers and thus the amount of computation as well as the size of the deployed model slightly increases compared to full fine-tuning. 

\noindent \textbf{Maintainability and Extensibility:} 
We find that rather than adding a new adapter for a new task, using MTL, where we combine the new task's data with 10\% of the previous tasks' data, not only gives a comparable performance but is also computationally comparable while benefitting from the cross-task transfer. 
As MTL only needs a new task specific classification head, it can be an excellent maintainable and extensible alternative to adapters. 

\noindent \textbf{Task Performance.} We show that both adapters and MTL can achieve comparable performance to full fine-tuning in both in-language and cross-language zero-shot settings.  
Our findings provide a realistic picture of adapters for NLU and show that while they are indeed parameter efficient, they suffer from compute limitations that can be addressed using approaches like MTL. We hope that our observations will spur further investigations into adapters and help in the development of PEFT approaches addressing the existing limitations of adapters. 
\section{Related Work}

\noindent \textbf{Parameter Efficient Fine-Tuning (PEFT):} \citet{zoph-etal-2016-transfer} was one of the earliest to work on PEFT by showing that fine-tuning a part of a pre-trained model reduces memory requirements and helps to avoid overfitting. Despite its simplicity, determining what part of the model should be fine-tuned involves exhaustive searching. However, this has spurred research into injecting fine-tunable components into the pre-trained model, the most prominent being works on Adapters \cite{pmlr-v97-houlsby19a,bapna-firat-2019-simple,DBLP:conf/iclr/HuSWALWWC22} which are tiny feed-forward layers injected after the self-attention and/or feed-forward layers of Transformer models \cite{NIPS2017_3f5ee243}. Learnable prompts \cite{li-liang-2021-prefix}, which are parameters appended to the key and values of the attention layers, can also be considered as adapters via a simple reformulation \cite{he2022towards}. Works such as compacters \cite{DBLP:conf/nips/MahabadiHR21} and IA$^{3}$ \cite{DBLP:journals/corr/abs-2205-05638} further focus on reducing the size of adapters. On the other hand, works on AdapterFusion \cite{pfeiffer-etal-2021-adapterfusion}, and MAD-X \cite{pfeiffer-etal-2020-mad} focus more on the transfer learning capabilities of adapters. However, these works mainly focus on parameter efficiency and leave out other aspects of efficiency, such as training time, deployability, maintainability, and cross-lingual transfer effectiveness. AdapterDrop \cite{ruckle-etal-2021-adapterdrop} proposes to reduce adapter training time but ignores other aforementioned aspects, a gap which we fill in this paper. While our study represents the empirical comparison of In-langauge, zero-shot performance and convergence time of PEFT method, multi-task learning, and fine-tuning methods, prior research has examined the instability of PEFT method. \citet{chen-etal-2022-revisiting} demonstrated the instability of PEFT in relation to weight initialization, training time, and training data order. They also compared the performance of PEFT and fine-tuning methods with respect to different dataset sizes. Following the broken protocol\footnote{dev set is used for early stopping as well as for reporting accuracy} issue as mentioned in the paper \cite[Section2]{chen-etal-2022-revisiting}, we have used different dev and test set for all our experiments. In addition to focusing on the observation that fine-tuning cannot be fully replaced by PEFT, our study has also demonstrated that multi-task learning can be an alternative to the PEFT method.

\noindent \textbf{Multilingual Pre-trained Models:} Ever since the introduction of BERT \cite{devlin-etal-2019-bert}, which is a pre-trained model which leverages monolingual data, there has been a steep improvement in the performance of downstream NLP tasks such as sentiment analysis, question answering and natural language inference. This was followed by massively multilingual pre-trained models such as the language group agnostic model XLM-R \cite{conneau-etal-2020-unsupervised}, and language group specific models IndicBERT \cite{DBLP:journals/corr/abs-2212-05409,kakwani-etal-2020-indicnlpsuite}, IndoBERT \cite{koto-etal-2020-indolem}, AfriBerta \cite{ogueji-etal-2021-small}, etc. Multilingual models enable cross-lingual transfer, allowing models to be fine-tuned on one language and be evalauted in a zero-shot on other languages. The efficiency of transfer via fine-tuning has not received due attention, and our work focuses on this aspect both in full fine-tuning and PEFT paradigms.

\noindent \textbf{Multi-Task Learning (MTL):} MTL focuses on fully-fine tuning one model for multiple tasks \cite{DBLP:conf/icml/Caruana93} but has only recently seen significant adoption \cite{wei2021finetuned,muennighoff2022crosslingual}. MTL benefits from cross-task transfer, which we also analyzed in this paper ($\S$\ref{sec:maintainability}). A general overview of MTL in deep learning can be found in \citet{DBLP:journals/corr/Ruder17a} and \citet{ DBLP:journals/corr/abs-2204-03508}. 

\begin{table*}[]
\centering
\small
\begin{tabular}{lllcccc}
\toprule
\textbf{Task Category} & \textbf{Train Data} & \textbf{Test Data} & \textbf{|Train|} & \textbf{|Test|} & \textbf{|Lang|} & \textbf{Metric} \\
\midrule
\multirow{4}{*}{\begin{tabular}[c]{@{}l@{}}\textbf{Sentence} \\ \textbf{Classification}\end{tabular}} & \begin{tabular}[c]{@{}l@{}}Amazon Multi Reviews\end{tabular} & IndicSentiment & 160k & 1000 & 11 & Acc. \\
\cmidrule(lr){2-7}
 & MultiNLI & IndicXNLI & 392k & 5000 & 11 & Acc. \\
 \cmidrule(lr){2-7}
 & SocialIQA & IndicCOPA & 33k & 500 & 11 & Acc. \\
 \cmidrule(lr){2-7}
 & PAWS & IndicParaphrase & 49k & 2002 & 10 & Acc. \\
 \midrule
\begin{tabular}[c]{@{}l@{}}\textbf{Token Classification}\end{tabular} & CoNLL-2003 & Naamapadam & 11k & 607-1080 & 11 & F1 \\
\midrule
\begin{tabular}[c]{@{}l@{}}\textbf{Question  Answering}\end{tabular} & SQuAD & IndicQA & 87k & 1517-2017 & 11 & F1 \\
\midrule
\end{tabular}
\caption{A summary of the tasks and datasets used. |Test| denotes the size of Test Data. |Train| is the size of English training sets. |Lang| denotes the number of languages for which we have evaluated its cross-lingual performance.}
\label{tab:tasks}
\end{table*}

\section{Experimental Setup}
\label{sec:exp_setup}

We now describe the fine-tuning approaches, tasks, datasets, languages, pre-trained models, and training settings.


\subsection{Fine-Tuning Methodologies}
\label{subsec:baselines}
Following are the fine-tuning approaches we experiment with.

\subsubsection{Non-Adapter Approaches} 
\noindent \textbf{Full Fine-Tuning} \cite{devlin-etal-2019-bert} is the standard approach, where all parameters are updated.

\noindent \textbf{Multi-Task Learning} \cite{liu-etal-2019-multi} is similar to full fine-tuning, except that it uses a shared encoder for all tasks, with each task having a task-specific ``head''.

\subsubsection{Adapter Approaches}
\noindent \textbf{Houlsby Adapter} \cite{pmlr-v97-houlsby19a} involves insertion of additional bottleneck feed-forward layers, after the self-attention and FFN sub-layers. We experiment with both, sequential and parallel (Houlsby sequential and Houlsby parallel) adapters \cite{he2022towards} .

\noindent \textbf{Bapna Adapter \textnormal{\cite{bapna-firat-2019-simple}}} inserts adapters only after FFN sub-layer. We again use both the sequential and parallel versions (Bapna sequential and Bapna parallel).

\noindent \textbf{LoRA \textnormal{\cite{DBLP:conf/iclr/HuSWALWWC22}} }inserts trainable low-rank matrices for the query and value matrices in the self-attention block to approximate the weight updates. 

\noindent \textbf{Compacter \textnormal{\cite{DBLP:conf/nips/MahabadiHR21}}} adapts the weights of neural networks using compact low-rank hypercomplex adapter layers. 

\noindent \textbf{Prefix-Tuning \textnormal{\cite{li-liang-2021-prefix}}} is inspired from textual prefixes. Here, $k$ trainable prefix vectors are prepended to the Keys (K) and values (V) in the self-attention block. 

\noindent \textbf{MAD-X \textnormal{\cite{pfeiffer-etal-2020-mad}}} is a method for cross-lingual transfer learning that pre-trains language-specific adapters for cross-lingual testing and task-specific adapters for the target task. 
%

\noindent \textbf{AdapterFusion \textnormal{\cite{pfeiffer-etal-2021-adapterfusion}}} uses adapters trained on other tasks for transfer learning as additional layers in the model for the downstream task. The fused layer is trained for the target task. 

\noindent \textbf{AdapterDrop \textnormal{\cite{ruckle-etal-2021-adapterdrop}}} aims to reduce the computational cost of training adapters by randomly dropping a subset of the adapters during each training iteration. 


While LoRA and prefix-tuning are not originally considered as adapters, \citet{he2022towards} have shown that they can be reformulated as adapters and thus all PEFT approaches we study in this paper are essentially adapters.

\subsection{Tasks, Datasets and Languages} 
\label{subsec:tasks}
We focus on 6 cross-lingual natural language understanding tasks from the IndicXTREME benchmark \cite{DBLP:journals/corr/abs-2212-05409} spanning 18 languages from 4 language families. These tasks can be broadly classified into sentence classification (4), token classification (1), and question answering (1). We give an overview in Table~\ref{tab:tasks}, including corpora sizes and metrics (Accuracy or F1) used for evaluation. Unless explicitly mentioned, we only train and validate on English data and evaluate on English test sets (supervised/in-language) as well as Indian language test sets in IndicXTREME (zero-shot). Please refer to Appendix~\ref{sec:appendix_tasklanguagesdetails} for details of tasks and languages.

\subsection{Pre-Trained Models}
\label{subsec:ft_models}
We mainly experiment with IndicBERT v2 \cite{DBLP:journals/corr/abs-2212-05409}  which is trained on the IndicCorp v2 corpus and supports 23 Indian languages and English. It is trained with the Masked Language Modeling (MLM) \cite{devlin-etal-2019-bert} objective. 
We also perform ablations with the \texttt{BASE} and \texttt{LARGE} versions of XLM-R \cite{conneau-etal-2020-unsupervised} on the chosen subset of languages. 

\noindent \textbf{Pretraining MAD-X} language adapter is done using the IndicCorp v2 \cite{DBLP:journals/corr/abs-2212-05409} dataset with MLM objective for the 11 Indic languages and English with 6.5M sentences sampled per language.

\subsection{Training Details}
\label{subsec:train_setup}
All models are trained with Adapter-hub \cite{pfeiffer-etal-2020-adapterhub}. 
All experiments are performed on Nvidia A100-SXM4 40GB GPUs and the results are reported by doing single run. We use the recommended/default settings in Adapter-hub but wherever possible, we performed hyperparameter tuning on the development set to determine optimal hyperparameters. Table~\ref{table:hyperparam-method} gives the search space and best performing hyperparameters for Houlsby, Bapna, LoRA and Prefix-Tuning. 
For MAD-X,  we have used the default configuration as in Adapter-hub for both language and task adapters, as shown in Table~\ref{table:hyperparam-method}.
For Adapter-fusion we have trained each task adapter in ST-A (single task adapter) style \cite{pfeiffer-etal-2021-adapterfusion}.

For all the tasks using the IndicBERT model, we train models for a maximum of 50 epochs with an early stopping patience of 3 epochs. We use 2,000 warmup steps for all tasks and settings, except for MTL, where we use 20,000 warmup steps due to the increased size of the training data. For a fair comparison across all settings, we use a batch size of 32 examples with a learning rate of 3e-5 and weight decay of 0.1. For MTL, we found that a weight decay of 0.01 gave the best results. For all the experiments FLOs reported are provided by the HF transformers library \cite{wolf-etal-2020-transformers}. 

\begin{table}[]
\small
\centering
\resizebox{\columnwidth}{!}{
\begin{tabular}{lcc}
\toprule
\textbf{Method} & \textbf{Hyperparameter} & \textbf{Search Space} \\
\midrule
Houlsby & $r=16$ & $r={2, 4, 8, 16}$ \\
Bapna & $r=16$ & $r={2, 4, 8, 16}$ \\
LoRA & $r = 8$, $\alpha = 16$ & $r={2, 4, 8, 16}$ \\
Prefix-Tuning & $l=30$ & $l={10, 20, 30, 40, 50}$\\
\bottomrule
\end{tabular}
}
\caption{This table reports the optimal reduction factor (r), prefix length (l) and LoRA $\alpha$ we have set for adapters. For those not listed in this table, we have used the default AdapterHub configurations.} 
\label{table:hyperparam-method}
\end{table}

\begingroup
\setlength{\tabcolsep}{3pt} 
\renewcommand{\arraystretch}{1} 
\begin{table*}[t]
\small
\centering
\begin{tabular}{cl|ccccccc|ccc}
\midrule
\#&\textbf{Method} & \begin{tabular}[c]{@{}c@{}}\textbf{AMR}\end{tabular} & \textbf{XNLI} & \textbf{COPA} & \textbf{PAWS} & \begin{tabular}[c]{@{}c@{}}\textbf{CoNLL}\\\textbf{2003}\end{tabular} & \textbf{SQuAD} & \textbf{Avg.} &  \begin{tabular}[c]{@{}c@{}}\textbf{\% $\uparrow$} \\\textbf{FLOs}\end{tabular} &\begin{tabular}[c]{@{}c@{}} \textbf{\% $\uparrow$} \\\textbf{Inference}\\ \textbf{time}\end{tabular} &  \begin{tabular}[c]{@{}c@{}}\textbf{\% $\uparrow$} \\ \textbf{\#Param.} \end{tabular}  \\\midrule
1 & Houlsby & \textbf{94.0} & 82.4 & 61.5 & 92.3 & 91.5 & 81.7 & 83.9 & 311.7 & 44.0 & 0.9 \\
2 & Bapna & 93.3 & 81.9 & 59.9 & 91.4 & 91.0 & 80.9 & 83.1 & 264.7 & 28.3 & 0.5 \\
3 & Houlsby Parallel & 93.1 & 82.5 & 61.4 & 90.6 & 92.2 & 82.0 & 83.6 & 185.1 & 41.5 & 0.9 \\
4 & Bapna Parallel & 93.1 & 82.7 & 60.5 & 91.3 & 91.1 & 81.4 & 83.4 & 199.9 & 21.2 & 0.5 \\
5 & Prefix Tuning & 93.8 & 82.6 & 61.1 & 92.2 & 91.5 & 81.0 & 83.7 & 186.5 & 33.8 & 3.8 \\
6 & LoRA & 93.4 & 80.3 & 57.4 & 90.2 & 90.4 & 79.5 & 81.8 & 226.2 & 23.1 & 0.3 \\
7 & Compacter & 92.8 & 74.8 & 50.8 & 72.7 & 89.2 & 73.0 & 75.5 & 371.4 & 100.5 & 0.2 \\
8 & Adapter Drop & 92.7 & 80.6 & 52.3 & 75.0 & 90.4 & 70.7 & 77.0 & 97.6 & 27.5 & 0.7 \\
9 & Adapter Fusion & 93.2 & 79.9 & 59.9 & 92.2 & 92.0 & 81.9 & 83.2 & 492.5 & 178.1 & 7.9 \\
10 & MAD-X - en & 93.6 & 82.1 & 56.9 & 91.0 & 91.5 & 81.1 & 82.7 & 1042.5 & 56.6 & 1.1 \\
11 & MAD-X - hi & 93.0 & 79.3 & 58.4 & 90.6 & 91.1 & 79.4 & 82.0 & 1025.7 & 56.6 & 1.1 \\
\midrule
&Best Adapter \# & 1 & 4 & 1 & 1 & 3 & 3 & 1 & 8 & 2 & 7 \\
\midrule
12 & FT & 93.8 & \textbf{83.0} & \textbf{62.3} & \textbf{93.0} & \textbf{92.8} & \textbf{82.1} & \textbf{84.5} & - & - & - \\
13 & MTL & 93.5 & 80.9 & 61.4 & 91.5 & 91.0 & \textbf{82.1} & 83.4 & 20.2 & 0.0 & 0.0 \\
\midrule
\multicolumn{1}{l}{} & Best method \# & 1 & 12 & 12 & 12 & 12 & 12, 13 & 12 & 12 & 12,13 & 12,13\\
\bottomrule
\end{tabular}
\caption{Comparison on \textbf{in-language (train and test on English)} performance of  FT and adapters for IndicBERT. We report F1 scores for CoNLL-2003 \& SQuAD, and accuracy for the other tasks. The abbreviation "AMR" refers to the Amazon Multilingual Review Dataset. The last three columns show the percent increase in FLOs, inference time, and the number of fine-tuned parameters compared to full fine-tuning respectively. Here, "best method \# " reports the best performing row for the respective task and "best adapter \# " reports the best performing adapter for the respective task.}
\label{table:en-indic-bert}
\end{table*}
\endgroup
\begin{table*}[t]
\small
\centering
\begin{tabular}{lccccccc}
\toprule
\textbf{Method} & \textbf{Sentiment} &\textbf{XNLI} & \textbf{COPA}& \textbf{Paraphrase} & \textbf{NER} & \textbf{QA} & \textbf{Total} \\
\midrule
Houlsby & 249.8 & 208.5 & 376.6 & 88.5 & 19.8 & 599.0 & 311.7 \\
Bapna & 185.2 & 246.5 & 321.0 & 43.6 & 77.7 & 456.7 & 264.7 \\
Houlsby Parallel & 105.3 & 208.5 & 274.2 & -5.8 & 88.0 & 275.0 & 185.1 \\
Bapna Parallel & 62.9 & 205.4 & 185.7 & 52.6 & 26.9 & 389.4 & 199.9 \\
Prefix Tuning & 190.9 & 237.2 & 179.4 & 96.2 & 77.4 & 198.1 & 186.5 \\
Lora & 223.2 & 203.1 & 168.0 & 93.6 & 143.6 & 402.9 & 226.2 \\
Compacter & 363.9 & 121.7 & 650.9 & 25.9 & 252.4 & 735.6 & 371.4 \\
Adapter Drop & 124.3 & 225.6 & 136.4 & -40.6 & 19.8 & 1.9 & 97.6 \\

\bottomrule
\end{tabular}
\caption{This table reports percentage increase of FLOs for several adapters across tasks with respect to full fine-tuning on IndicBert model. Column "Total" reports the percentage increase in total FLOs for each method with respect to full fine-tuning (FLOs are added across all tasks). Since, for Adapter Fusion and MAD-X, task adapters and language adapters, respectively, are shared across tasks, training FLOs are also shared across tasks. Thus, for these two approaches, FLOs cannot be reported accurately for individual tasks.}
\label{compute-percent: indicbert}
\end{table*}

\begin{table*}[]
\small
\centering
\begin{tabular}{clccccccc}
\toprule
\#&\textbf{Method} & \begin{tabular}[c]{@{}c@{}}\textbf{Indic}\\ \textbf{Sentiment}\end{tabular} & \begin{tabular}[c]{@{}c@{}}\textbf{Indic}\\\textbf{XNLI}\end{tabular} & \begin{tabular}[c]{@{}c@{}}\textbf{Indic}\\ \textbf{COPA}\end{tabular} & \begin{tabular}[c]{@{}c@{}}\textbf{Indic}\\ \textbf{XPara}\end{tabular} & \begin{tabular}[c]{@{}c@{}}\textbf{Naama-}\\ \textbf{padam}\end{tabular} & \textbf{IndicQA} & \textbf{Avg.} \\ \midrule
1 & Houlsby & 89.7 & 72.9 & 64.1 & 57.4 & 65.5 & \textbf{50.0} & 66.6  \\
2 & Bapna & 89.0 & 72.1 & 60.9 & 55.9 & 65.2 & 48.6 & 65.3  \\
3 & Houlsby Parallel & 90.3 & 72.4 & 63.7 & 55.8 & 66.7 & 49.2 & 66.4  \\
4 & Bapna Parallel & 89.9 & 72.5 & 61.4 & 56.3 & 64.7 & 48.9 & 65.6  \\
5 & Prefix Tuning & 88.2 & \textbf{73.5} & \textbf{65.3} & 55.8 & \textbf{67.1} & 48.4 & 66.4  \\
6 & Lora & 85.7 & 70.7 & 60.7 & 55.0 & 63.3 & 47.4 & 63.8\\
7 & Compacter & 88.5 & 69.9 & 63.2 & 50.8 & 61.3 & 46.4 & 63.4  \\
8 & Adapter Drop & 87.8 & 72.0 & 61.8 & 52.9 & 64.4 & 44.4 & 63.9  \\
9 & Adapter Fusion & 89.3 & 70.8 & 59.3 & 56.3 & 66.9 & 48.7 & 65.2 \\
10 & MAD-X - en & 89.6 & 72.4 & 62.6 & 55.9 & 66.0 & 47.6 & 65.7  \\
11 & MAD-X - hi & 88.6 & 70.8 & 63.1 & 56.5 & 64.1 & 47.4 & 65.1  \\
\midrule
& Best Adapter \# & 3 & 5 & 5 & 1 & 5 & 1 & 1\\
\midrule
12 & FT & \textbf{90.9} & 72.9 & 62.5 & 57.3 & 66.7 & 49.3 & 66.6 \\
13 & MTL & 90.2 & 70.7 & \textbf{65.3} & \textbf{74.3} & 65.3 & 45.5 & \textbf{68.6} \\
\midrule
\multicolumn{1}{l}{} & Best method \# & 12 & 5 & 5, 13 & 13 & 5 & 1 & 13 \\
\bottomrule
\end{tabular}
\caption{Comparison on \textbf{cross-lingual (train on English test on Indic)} performance of FT and adapters for IndicBERT. We report F1 scores for Naamapadam \& IndicQA, and accuracy for the other tasks. Here, "best method \# " reports the best performing row for the respective task and "best adapter \# " reports the best performing adapter for the respective task.}
\label{table:avg-xl-indicbert}
\end{table*}

\section{Results}
We now report results comparing various efficiency aspects of adapter and non-adapter approaches. Tables \ref{table:en-indic-bert} and \ref{table:avg-xl-indicbert} respectively show the in-language (train and test on English) and cross-lingual (train on English and test on Indic) results averaged across Indic languages. See Appendix \ref{sec:appendix_tasklevelsensitivity} for per-language performances. We present our key observations in the following sub-sections.

\subsection{Parameter Efficiency}
\label{sec:param-efficiency}
\noindent \textbf{Adapters are parameter-efficient, but no single adapter is best:} It is clear that there is no single adapter that performs best in all the tasks. This observation holds true in both in-language and cross-lingual settings, where one method performs best in the in-language setting but might not be the best in the cross-lingual setting. 
Compacter and LORA consistently give the lowest performance, possibly due to the small number of parameters they fine-tune (they add only 0.2\% - 0.3\% tunable parameters to the model). On the other hand, Adapter Fusion, Prefix Tuning, and MADX add between 1.1\% to 7.9\% tunable parameters but still perform poorly as compared to the Houlsby adapter, which only adds 0.9\% parameters. In general, we recommend the Houlsby adapter as it tends to perform well across multiple tasks and languages on average.


\subsection{Compute Efficiency}
\label{sec:train-efficiency}
\noindent We calculated the total number of FLOs for all methods for all tasks and report percentage increases relative to full fine-tuning in Table~\ref{compute-percent: indicbert}. Task specific details of model convergence and absolute FLOs (Tables~\ref{compute-avg-absolute: indicbert})  are available in the Appendix.


\noindent \textbf{Full fine-tuning} is the fastest by a significant margin. While \textbf{adapters} methods are \textit{parameter} efficient, they are not computationally efficient when fine-tuning. In practice, they consume more FLOs to converge and achieve performance comparable to full fine-tuning. AdapterDrop (row 8 in Table~\ref{table:en-indic-bert}) exhibits the least increase in FLOs (97.6\%) but also suffers from reduced performance. MAD-X (rows 10, 11) is the costliest (1042.5\%-1025.7\%) in terms of FLOs but still gives poor results compared to full fine-tuning. The best performing adapter (Houlsby, row 1) is also computationally very expensive. These results clearly show that adapters are computationally very costly while achieving comparable or worse performance compared to full fine-tuning.


\noindent \textbf{MTL} is a cost-efficient alternative to adapters given that it only uses 20\% more FLOs than full fine-tuning while achieving performance comparable to the best adapter (Houlsby gives 83.9\% \& MTL gives 83.4\%). Further, MTL exhibits the best average cross-lingual performance with respect to adapters as well as full fine-tuning. It should be noted that MTL significantly benefits the paraphrasing task via cross-task transfer, exhibiting a performance increase of 16.9\% accuracy over full fine-tuning in a cross-lingual setting (experiments in further sections show that paraphrasing benefits from the NLI task). Thus, if the full set of tasks to be supported is known \textit{a priori}, MTL is simpler and equivalent to adapters in downstream performance, while being more cost-efficient. \citet{sanh2022multitask} show that MTL enables zero-shot task generalization, further enhancing the attractiveness of MTL over adapters.

\subsection{Inference Overhead}
\label{sec:inference-deployability}
 Table~\ref{table:en-indic-bert} also shows the increase in inference time for different approaches compared to full fine-tuning. MTL does not add any overhead over full fine-tuning since no new parameters are added to the model. On the other hand, adapters have a non-trivial overhead in inference time due to additional parameters. The Bapna parallel and LoRA methods show least increase in inference time (of 21.2\% and 23.1\%, respectively), since they are parallel adapters. Bapna parallel has lesser inference time than Houlsby parallel as it has almost half the number of parameters. The adapter fusion method has the highest inference time as it combines all six task adapters and has an additional fused layer. It also has the maximum number of additional parameters. Although Compacter has the least number of parameters, its inference time is 100.5\% more than fine-tuning because the compact low-rank hypercomplex weight matrices are converted to high-rank ones via the Kronecker product. These high-rank matrices are actually used during the forward pass and this two-step process slows down inferencing\footnote{The current implementation does not pre-compute the high-rank matrices and thus there is a possibility of reducing the inference time of Compacter, although it will not be faster than the Houlsby adapter to which it is architecturally similar.}. 

\begingroup
\setlength{\tabcolsep}{2pt} 
\renewcommand{\arraystretch}{1} 
\begin{table*}[h!]
\small
\centering
\begin{tabular}{llccccccccc}
\toprule
\begin{tabular}[c]{@{}c@{}}\textbf{Target}\\ \textbf{Task}\end{tabular} & \textbf{Step} & \begin{tabular}[c]{@{}c@{}}\textbf{Indic}\\ \textbf{Sentiment}\end{tabular} & \begin{tabular}[c]{@{}c@{}}\textbf{Indic}\\ \textbf{XNLI}\end{tabular} & \begin{tabular}[c]{@{}c@{}}\textbf{Indic}\\ \textbf{COPA}\end{tabular} & \begin{tabular}[c]{@{}c@{}}\textbf{Indic}\\ \textbf{XPara}\end{tabular} & \begin{tabular}[c]{@{}c@{}}\textbf{Naama-}\\ \textbf{padam}\end{tabular} & \textbf{IndicQA} & \textbf{Avg -1} & \textbf{Avg} & \begin{tabular}[c]{@{}c@{}}\textbf{\% $\uparrow$} \\\textbf{FLOs}\end{tabular}\\
\midrule
 Baseline & Full FT & \textbf{90.9} & \textbf{72.9} & 62.5 & 57.3 & 66.7 & 49.3 & - & 66.6&- \\
 & MTL & 88.5 & 71.2 & 64.9 & 74.0 & 65.8 & 45.4 & - & 68.3 &20.2 \\
\midrule
 Best Adapter & Houlsby & 89.7 & \textbf{72.9} & 64.1 & 57.4 & 65.5 & \textbf{50.0} &-& 66.6 &311.7 \\
\midrule
Sentiment & MTL$_{-1}$ & - & 71.5 & \cellcolor{blue!25}64.8 & \cellcolor{blue!25}74.8 & 65.1 & 46.9 & 64.6 & -  & - \\

 & {MTL}$_{+tgt+old}$ & \cellcolor{blue!25}90.2 & 70.8 & 61.9 & \cellcolor{blue!25}72.9 & \cellcolor{blue!25}66.1 & 48.8 & 64.1 & \cellcolor{blue!25}\textbf{68.4} & 2.3\\
 
 & MTL$_{+tgt}$ & 89.1 & 54.9 & 52.2 & \cellcolor{blue!25}67.3 & 40.4 & 34.9 & 50.0 & 56.5& 1.7\\
\midrule
XNLI & MTL$_{-1}$ & \cellcolor{blue!25}90.5 & - & \cellcolor{blue!25}\textbf{67.8} & 56.7 & 63.6 & 47.4 & 65.2 & -& -  \\

 & {MTL}$_{+tgt+old}$ & \cellcolor{blue!25}90.8 & 71.2 & 63.6 & \cellcolor{blue!25}68.7 & 59.6 & 48.3 & 66.2 & \cellcolor{blue!25}67.0 & 20.5\\
 
 &MTL$_{+tgt}$ & 86.0 & 70.5 & \cellcolor{blue!25}64.5 & \cellcolor{blue!25}73.3 & 56.2 & 15.1 & 59.0 & 60.9 & 12.1 \\
\midrule
COPA & MTL$_{-1}$ & 88.8 & 72.3 & - & \cellcolor{blue!25}73.7 & 65.0 & 48.4 & 69.7 & - & -\\

 & {MTL}$_{+tgt+old}$ & 88.3 & 69.7 & \cellcolor{blue!25}65.6 & \cellcolor{blue!25}74.8 & 65.4 & 43.9 & 68.4 & \cellcolor{blue!25}67.9  & 15.3\\
 
 & MTL$_{+tgt}$ & 89.5 & 66.4 & \cellcolor{blue!25}66.0 & \cellcolor{blue!25}75.5 & 62.7 & 46.4 & 68.1 & \cellcolor{blue!25}67.7 & 9.3 \\
\midrule
Paraphrase & MTL$_{-1}$ & 86.0 & 70.2 & \cellcolor{blue!25}64.4 & - & 65.0 & 45.0 & 66.1 &  -&-  \\

 & {MTL}$_{+tgt+old}$ & 87.4 & 69.8 & \cellcolor{blue!25}64.2 & \cellcolor{blue!25}\textbf{77.8} & 65.0 & 45.4 & 66.4 & \cellcolor{blue!25}68.3 & 32.4 \\
 
 &  MTL$_{+tgt}$ & 81.1 & 66.0 & \cellcolor{blue!25}64.4 & \cellcolor{blue!25}73.1 & 30.1 & 42.5 & 56.8 & 59.5  & 24.1\\
\midrule
NER & MTL$_{-1}$ & 88.0 & 72.5 & \cellcolor{blue!25}65.7 & \cellcolor{blue!25}77.3 & - & 47.7 & 70.3 & - &  - \\

 & {MTL}$_{+tgt+old}$ & 86.7 & 71.2 & \cellcolor{blue!25}64.4 & \cellcolor{blue!25}76.3 & 65.2 & 45.1 & 68.7 & \cellcolor{blue!25}68.1  & 68.2 \\
 
 &  MTL$_{+tgt}$ & 83.8 & 67.9 & 62.3 & 57.3 & \cellcolor{blue!25}\textbf{68.5} & 39.8 & 62.2 & 63.2  & 59.8\\
\midrule
QA & MTL$_{-1}$ & 89.2 & 72.3 & \cellcolor{blue!25}64.9 & \cellcolor{blue!25}74.9 & 65.4 & - & 73.3 & - & - \\

 & {MTL}$_{+tgt+old}$ & 85.9 & 71.1 & 63.9 & \cellcolor{blue!25}75.7 & 62.3 & 46.8 & 71.8 & \cellcolor{blue!25}67.6  & 25.8\\
 
 & MTL$_{+tgt}$ & 84.9 & 68.2 & \cellcolor{blue!25}65.9 & \cellcolor{blue!25}66.9 & 23.7 & 46.6 & 61.9 & 59.4& 21.2 \\
 \bottomrule
\end{tabular}
\caption{This table reports \textbf{cross-lingual (train on English test on Indic)} performance for maintainability of MTL. "Target task" is held out task i.e. pre-trained IndicBERT model is fine-tuned on the remaining 5 task representing  MTL$_{-1}$ model. {MTL}$_{+tgt+old}$ represents continual fine-tuning of the MTL$_{-1}$ model on the target task dataset and 10\% of the existing task dataset. {MTL}$_{+tgt}$ represents continual fine-tuning of the MTL$_{-1}$ model on the target task dataset. "Avg -1 " reports the cross-lingual performance averaged over the tasks included in MTL$_{-1}$ step. "Avg" reports the cross-lingual performance averaged over all 6 task. Here, column "\%$\uparrow$FLOs" reports the relative percent increase in the total computation cost for adding all 6 task to the model with respect to the total computation cost of fine-tuning. Here, text bold indicates the best value in the column and colored cell represent MTL is performing better than the Best Adapter method.}

\label{table:MTL_avg_xL}
\end{table*}
\endgroup

\begin{table*}[]
\small
\centering
\begin{tabular}{l|ccccc|ccccc}
\toprule
\textbf{XL} & \multicolumn{5}{c}{\textbf{XLMR-Base}} & \multicolumn{5}{|c}{\textbf{XLMR-Large}} \\
\midrule
\textbf{Method} & \textbf{NER} & \textbf{XNLI} & \textbf{QA} & \textbf{Avg.} & \textbf{\%$\uparrow$FLOs} & \textbf{NER} & \textbf{XNLI} & \textbf{QA} & \textbf{Avg.} & \textbf{\%$\uparrow$FLOs}\\
\midrule
Houlsby & 61.0 & 72.6 & \textbf{72.5} & \textbf{68.7} & 484.3 & \textbf{64.6} & 76.2 & \textbf{78.6} & 73.1 & 200.5 \\
Bapna & 58.3 & 71.3 & 71.0 & 66.8 & 547.1 & 64.3 & 76.7 & 78.0 & 73.0 & 139.9 \\
Houlsby parallel & 59.2 & 72.8 & 71.2 & 67.8 & 197.0 & 65.3 & 78.7 & 77.8 & 73.9 & 143.3 \\
Bapna parallel & 57.1 & 70.3 & 69.7 & 65.7 & 409.1 & 63.1 & \textbf{78.8} & 77.6 & 73.2 & 168.6 \\
Prefixtuning & 58.5 & 69.9 & 68.8 & 65.7 & 256.5 & 64.7 & 78.7 & 77.6 & 73.7 & 287.3 \\
LORA & 58.6 & 70.5 & 68.4 & 65.8 & 734.7 & 62.3 & 76.9 & 77.1 & 72.1 & 270.0 \\
Compacter & 55.1 & 66.8 & 64.1 & 62.0 & 805.3 & 58.5 & 76.4 & 75.3 & 70.1 & 490.1 \\
Adapter drop & 60.5 & 70.2 & 71.3 & 67.3 & 345.1 & 64.6 & \textbf{78.8} & 78.5 & \textbf{74.0} & 214.1 \\
FT & \textbf{61.7} & \textbf{73.7} & 70.8 & \textbf{68.7} & - & 63.9 & 77.0 & 78.0 & 73.0 & - \\
\bottomrule
\end{tabular}
\caption{Comparison on \textbf{cross-lingual} performance of FT and adapters for XLMR-Base and XLMR-Large model. "Avg." reports the average cross-lingual perfromance across all task. "\%$\uparrow$FLOs" reports the relative increase in FLOs with respect to fine-tuning.}
\label{Overall XL: XLMR-computation}
\end{table*}

\subsection{Maintainability and Extensibility}
\label{sec:maintainability}

The primary advantage of adapters is the ability to `plug-and-play' modules, thus making it easy to extend a pre-trained model to new tasks without having to make a copy for the new task or impacting performance on other tasks. This reduces memory requirements at inference time and makes the system more modular, maintainable and extensible. We have already seen that MTL models offer the same performance with no additional parameters and at a lower computational cost compared to adapters. To see if they can also be easily extensible, we experiment with the following setup. 

We hold out one task (the \textit{target task}) and fine-tune the pre-trained model on the remaining tasks (resulting in model \textbf{MTL$_{-1}$}). Next, we continue fine-tuning the model on the target task  as well as 10\% data from the tasks the model has already seen. A sample from the older tasks is included in the fine-tuning mix to avoid catastrophic forgetting \cite{MCCLOSKEY1989109, FRENCH1999128}. For comparison, we also perform continued fine-tuning on the \textbf{target task only} (model: \textbf{MTL$_{+tgt}$}) as well as fine-tuning on all available tasks (model: \textbf{MTL}). 

The results of these experiments are shown in Table~\ref{table:MTL_avg_xL} for cross-lingual settings (and Table~\ref{table:MTL_en} in Appendix for in-language settings). We see that the target task's performance is comparable to both full fine-tuning and MTL with all tasks. Thus, new tasks can be added to an existing MTL model while retaining the same performance as full FT or MTL. Moreover, we see that the \textbf{MTL}$_{+tgt+old}$ model also retains performance for the older tasks. We also see that if sample data from the already supported tasks is not used, the model suffers from catastrophic forgetting (model: \textbf{MTL}$_{+tgt}$). Thus, a simple adaptation of MTL can support multiple tasks in an extensible manner. 

The fine-tuning computational cost for \textbf{MTL}$_{+tgt+old}$ is the sum of computational costs for (a) fine-tuning \textbf{MTL$_{-1}$} and (b) continued fine-tuning required to extend model for the target task. In Table~\ref{table:MTL_avg_xL}, column "\%$\uparrow$FLOs" reports the percentage increase in total FLOs(sum of (a) and (b)) with respect to total fine-tuning FLOs(i.e. Fine-tuning FLOs sum over all task).
As observed, holding out sentiment task, and then continual learning of sentiment task along with 10\% data of existing tasks takes only 2.3\% more relative FLOs. The maximum cost is taken by NER task with 68.2\% more relative FLOs. Holding out one task and then adding the held out task on an average takes 27.4\% more relative FLOs, while adding all tasks at once takes 20.2\% more relative FLOs. Nonetheless, this is still more cost-effective than the best-performing adapter methods. For instance, the Houlsby adapter requires around $~$311\% more computation compared to full fine-tuning. Thus, we see maintainability of MTL cost-effective. 
However, average cross-lingual performance for MTL maintainability (as shown in Table~\ref{table:MTL_avg_xL}), is slightly inflated due to the inclusion of the paraphrase task. If the average MTL performance is calculated without the paraphrase task (i.e. only considering the remaining five tasks), a slight decrease in performance is observed.

\subsection{Effect of Model Size}

To further study the effect of model size on different adapters, we experiment with two different pre-trained models trained on the same pretraining data but differing only in model size. Specifically, we compare the XLMR-base and XLMR-large models \cite{conneau-etal-2020-unsupervised} which have 270M and 550M parameters, respectively. We evaluate the adapters on the XNLI, XQuAD and NER tasks from the XTREME benchmark \cite{DBLP:journals/corr/abs-2003-11080}. We use the English dataset for training and test the cross-lingual zero-shot performance on 14 languages for XNLI and WikiANN and 11 languages for XQuAD. The results are shown in Table \ref{Overall XL: XLMR-computation}. We can see that as the model size increase, the adaptation time relative to full fine-tuning time reduces. Thus, for large language models, we might see a trend of adapters being increasingly cost-efficient. In fact, recent work on large language models have shown adapters to be promising \cite{DBLP:journals/corr/abs-2212-09535}. 
However, larger models still need heavy compute and deploying them is still challenging. In this case, there is a line of work that distills LLMs which can then be fine-tuned \cite{DBLP:journals/corr/abs-2110-04711}. Given that adapters do not have much compute efficiency in smaller models, full-fine tuning or MTL are excellent contenders. 

\subsection{Key Takeaway}
Fig~\ref{fig:cc-dist} shows a unified summary of task performance and fine-tuning compute required for the various approaches discussed in the paper. 
Summarizing observations previously discussed, we see that MTL outperforms or is comparable to all adapters in in-language and cross-language zero-shot settings (particularly for smaller models). Hence, we recommend that MTL should be considered as an alternative to adapters in constrained scenarios where relatively smaller models are preferred, computational budgets are limited and extensibility is important. 

\section{Conclusion}
In this paper, we have conducted a comprehensive analysis of adapters across different languages and tasks to evaluate their advantages in terms of training/deployment efficiency and maintainability/extensibility. We compared adapters with simpler baseline methods, including fine-tuning and multi-task learning, in supervised/in-language as well as zero-shot cross-lingual settings, and found that these simpler methods are more computationally efficient and have better deployment efficiency, while achieving the comparable performance as that of adapters. 
Additionally, we conducted extensive experiments to show that multi-task learning is a relatively more cost-effective alternative to the adapters in terms of maintainability, as it allows the model to be extended for new tasks at a lower cost than adapters.  Therefore, we suggest that simpler baselines be used for moderately sized models, as they are more efficient than adapters.

\section*{Acknowledgements}
We would like to thank the Ministry of Electronics and Information Technology\footnote{\url{https://www.meity.gov.in/}} of the Government of India for their generous grant through the Digital India Bhashini project\footnote{\url{https://www.bhashini.gov.in/}}. We also thank the Centre for Development of Advanced Computing\footnote{\url{ https://www.cdac.in/index.aspx?id=pune}} for providing compute time on the Param Siddhi Supercomputer. We also thank Nilekani Philanthropies for their generous grant towards building datasets, models, tools and resources for Indic languages. We also thank Microsoft for their grant to support research on Indic languages.

\section*{Limitations}
We identify the following limitations of our work:

\begin{itemize}
    \item Our study is limited to NLU and some of our observations might not apply in Natural Language Generation (NLG) settings. While for NLU cross-lingual transfer through full fine-tuning is as effective as adapters, in NLG full fine-tuning for zero-shot cross-lingual NLG is unreliable due to the risk of catastrophic forgetting. Therefore, adapters might be more important for NLG \cite{DBLP:journals/corr/abs-2205-12647}.
    
    \item We primarily focus on smaller pre-trained models because larger models require significant computing resources that not everyone may have access to, and therefore, our findings may not be applicable to larger models with billions of parameters. However, active research on compressing pre-trained models indicates that fine-tuning compact pre-trained models will remain a significant area of research.

    \item Our analysis focus on 6 NLU tasks, which is relatively fewer compared to the total number of tasks in benchmarks such as BIG-Bench \cite{srivastava2022imitation}. Although focusing on a larger number of tasks will increase the credibility of our studies, our focus on cross-lingual performance means that we are currently limited by the availability of benchmarking data in other languages for these large number of tasks.
\end{itemize}

\section*{Ethics Statement}
All of the datasets used in this study were publicly available, and no annotators were employed for data collection. We confirm that the datasets we used did not contain any harmful content. We have cited the datasets and relevant works used in this study.



\bibliography{custom, anthology}

\begin{thebibliography}{43}
\expandafter\ifx\csname natexlab\endcsname\relax\def\natexlab#1{#1}\fi

\bibitem[{Bapna and Firat(2019)}]{bapna-firat-2019-simple}
Ankur Bapna and Orhan Firat. 2019.
\newblock \href {https://doi.org/10.18653/v1/D19-1165} {Simple, scalable
  adaptation for neural machine translation}.
\newblock In \emph{Proceedings of the 2019 Conference on Empirical Methods in
  Natural Language Processing and the 9th International Joint Conference on
  Natural Language Processing (EMNLP-IJCNLP)}, pages 1538--1548, Hong Kong,
  China. Association for Computational Linguistics.

\bibitem[{Caruana(1993)}]{DBLP:conf/icml/Caruana93}
Rich Caruana. 1993.
\newblock \href {https://doi.org/10.1016/b978-1-55860-307-3.50012-5} {Multitask
  learning: {A} knowledge-based source of inductive bias}.
\newblock In \emph{Machine Learning, Proceedings of the Tenth International
  Conference, University of Massachusetts, Amherst, MA, USA, June 27-29, 1993},
  pages 41--48. Morgan Kaufmann.

\bibitem[{Chen et~al.(2022)Chen, Liu, Meng, and
  Liang}]{chen-etal-2022-revisiting}
Guanzheng Chen, Fangyu Liu, Zaiqiao Meng, and Shangsong Liang. 2022.
\newblock \href {https://aclanthology.org/2022.emnlp-main.168} {Revisiting
  parameter-efficient tuning: Are we really there yet?}
\newblock In \emph{Proceedings of the 2022 Conference on Empirical Methods in
  Natural Language Processing}, pages 2612--2626, Abu Dhabi, United Arab
  Emirates. Association for Computational Linguistics.

\bibitem[{Conneau et~al.(2020)Conneau, Khandelwal, Goyal, Chaudhary, Wenzek,
  Guzm{\'a}n, Grave, Ott, Zettlemoyer, and
  Stoyanov}]{conneau-etal-2020-unsupervised}
Alexis Conneau, Kartikay Khandelwal, Naman Goyal, Vishrav Chaudhary, Guillaume
  Wenzek, Francisco Guzm{\'a}n, Edouard Grave, Myle Ott, Luke Zettlemoyer, and
  Veselin Stoyanov. 2020.
\newblock \href {https://doi.org/10.18653/v1/2020.acl-main.747} {Unsupervised
  cross-lingual representation learning at scale}.
\newblock In \emph{Proceedings of the 58th Annual Meeting of the Association
  for Computational Linguistics}, pages 8440--8451, Online. Association for
  Computational Linguistics.

\bibitem[{Devlin et~al.(2019)Devlin, Chang, Lee, and
  Toutanova}]{devlin-etal-2019-bert}
Jacob Devlin, Ming-Wei Chang, Kenton Lee, and Kristina Toutanova. 2019.
\newblock \href {https://doi.org/10.18653/v1/N19-1423} {{BERT}: Pre-training of
  deep bidirectional transformers for language understanding}.
\newblock In \emph{Proceedings of the 2019 Conference of the North {A}merican
  Chapter of the Association for Computational Linguistics: Human Language
  Technologies, Volume 1 (Long and Short Papers)}, pages 4171--4186,
  Minneapolis, Minnesota. Association for Computational Linguistics.

\bibitem[{Doddapaneni et~al.(2022)Doddapaneni, Aralikatte, Ramesh, Goyal,
  Khapra, Kunchukuttan, and Kumar}]{DBLP:journals/corr/abs-2212-05409}
Sumanth Doddapaneni, Rahul Aralikatte, Gowtham Ramesh, Shreya Goyal, Mitesh~M.
  Khapra, Anoop Kunchukuttan, and Pratyush Kumar. 2022.
\newblock \href {https://doi.org/10.48550/arXiv.2212.05409} {Indicxtreme: {A}
  multi-task benchmark for evaluating indic languages}.
\newblock \emph{CoRR}, abs/2212.05409.

\bibitem[{French(1999)}]{FRENCH1999128}
Robert~M. French. 1999.
\newblock \href {https://doi.org/https://doi.org/10.1016/S1364-6613(99)01294-2}
  {Catastrophic forgetting in connectionist networks}.
\newblock \emph{Trends in Cognitive Sciences}, 3(4):128--135.

\bibitem[{Ganesan et~al.(2021)Ganesan, Ramesh, and
  Kumar}]{DBLP:journals/corr/abs-2110-04711}
Vinod Ganesan, Gowtham Ramesh, and Pratyush Kumar. 2021.
\newblock \href {http://arxiv.org/abs/2110.04711} {Supershaper: Task-agnostic
  super pre-training of {BERT} models with variable hidden dimensions}.
\newblock \emph{CoRR}, abs/2110.04711.

\bibitem[{He et~al.(2022)He, Zhou, Ma, Berg-Kirkpatrick, and
  Neubig}]{he2022towards}
Junxian He, Chunting Zhou, Xuezhe Ma, Taylor Berg-Kirkpatrick, and Graham
  Neubig. 2022.
\newblock \href {https://openreview.net/forum?id=0RDcd5Axok} {Towards a unified
  view of parameter-efficient transfer learning}.
\newblock In \emph{International Conference on Learning Representations}.

\bibitem[{Houlsby et~al.(2019)Houlsby, Giurgiu, Jastrzebski, Morrone,
  De~Laroussilhe, Gesmundo, Attariyan, and Gelly}]{pmlr-v97-houlsby19a}
Neil Houlsby, Andrei Giurgiu, Stanislaw Jastrzebski, Bruna Morrone, Quentin
  De~Laroussilhe, Andrea Gesmundo, Mona Attariyan, and Sylvain Gelly. 2019.
\newblock \href {https://proceedings.mlr.press/v97/houlsby19a.html}
  {Parameter-efficient transfer learning for {NLP}}.
\newblock In \emph{Proceedings of the 36th International Conference on Machine
  Learning}, volume~97 of \emph{Proceedings of Machine Learning Research},
  pages 2790--2799. PMLR.

\bibitem[{Hu et~al.(2022)Hu, Shen, Wallis, Allen{-}Zhu, Li, Wang, Wang, and
  Chen}]{DBLP:conf/iclr/HuSWALWWC22}
Edward~J. Hu, Yelong Shen, Phillip Wallis, Zeyuan Allen{-}Zhu, Yuanzhi Li,
  Shean Wang, Lu~Wang, and Weizhu Chen. 2022.
\newblock \href {https://openreview.net/forum?id=nZeVKeeFYf9} {Lora: Low-rank
  adaptation of large language models}.
\newblock In \emph{The Tenth International Conference on Learning
  Representations, {ICLR} 2022, Virtual Event, April 25-29, 2022}.
  OpenReview.net.

\bibitem[{Hu et~al.(2020)Hu, Ruder, Siddhant, Neubig, Firat, and
  Johnson}]{DBLP:journals/corr/abs-2003-11080}
Junjie Hu, Sebastian Ruder, Aditya Siddhant, Graham Neubig, Orhan Firat, and
  Melvin Johnson. 2020.
\newblock \href {http://arxiv.org/abs/2003.11080} {{XTREME:} {A} massively
  multilingual multi-task benchmark for evaluating cross-lingual
  generalization}.
\newblock \emph{CoRR}, abs/2003.11080.

\bibitem[{Kakwani et~al.(2020)Kakwani, Kunchukuttan, Golla, N.C.,
  Bhattacharyya, Khapra, and Kumar}]{kakwani-etal-2020-indicnlpsuite}
Divyanshu Kakwani, Anoop Kunchukuttan, Satish Golla, Gokul N.C., Avik
  Bhattacharyya, Mitesh~M. Khapra, and Pratyush Kumar. 2020.
\newblock \href {https://doi.org/10.18653/v1/2020.findings-emnlp.445}
  {{I}ndic{NLPS}uite: Monolingual corpora, evaluation benchmarks and
  pre-trained multilingual language models for {I}ndian languages}.
\newblock In \emph{Findings of the Association for Computational Linguistics:
  EMNLP 2020}, pages 4948--4961, Online. Association for Computational
  Linguistics.

\bibitem[{Keung et~al.(2020)Keung, Lu, Szarvas, and
  Smith}]{DBLP:conf/emnlp/KeungLSS20}
Phillip Keung, Yichao Lu, Gy{\"{o}}rgy Szarvas, and Noah~A. Smith. 2020.
\newblock \href {https://doi.org/10.18653/v1/2020.emnlp-main.369} {The
  multilingual amazon reviews corpus}.
\newblock In \emph{Proceedings of the 2020 Conference on Empirical Methods in
  Natural Language Processing, {EMNLP} 2020, Online, November 16-20, 2020},
  pages 4563--4568. Association for Computational Linguistics.

\bibitem[{Koto et~al.(2020)Koto, Rahimi, Lau, and
  Baldwin}]{koto-etal-2020-indolem}
Fajri Koto, Afshin Rahimi, Jey~Han Lau, and Timothy Baldwin. 2020.
\newblock \href {https://doi.org/10.18653/v1/2020.coling-main.66} {{I}ndo{LEM}
  and {I}ndo{BERT}: A benchmark dataset and pre-trained language model for
  {I}ndonesian {NLP}}.
\newblock In \emph{Proceedings of the 28th International Conference on
  Computational Linguistics}, pages 757--770, Barcelona, Spain (Online).
  International Committee on Computational Linguistics.

\bibitem[{Li and Liang(2021)}]{li-liang-2021-prefix}
Xiang~Lisa Li and Percy Liang. 2021.
\newblock \href {https://doi.org/10.18653/v1/2021.acl-long.353} {Prefix-tuning:
  Optimizing continuous prompts for generation}.
\newblock In \emph{Proceedings of the 59th Annual Meeting of the Association
  for Computational Linguistics and the 11th International Joint Conference on
  Natural Language Processing (Volume 1: Long Papers)}, pages 4582--4597,
  Online. Association for Computational Linguistics.

\bibitem[{Liu et~al.(2022)Liu, Tam, Muqeeth, Mohta, Huang, Bansal, and
  Raffel}]{DBLP:journals/corr/abs-2205-05638}
Haokun Liu, Derek Tam, Mohammed Muqeeth, Jay Mohta, Tenghao Huang, Mohit
  Bansal, and Colin Raffel. 2022.
\newblock \href {https://doi.org/10.48550/arXiv.2205.05638} {Few-shot
  parameter-efficient fine-tuning is better and cheaper than in-context
  learning}.
\newblock \emph{CoRR}, abs/2205.05638.

\bibitem[{Liu et~al.(2019{\natexlab{a}})Liu, He, Chen, and
  Gao}]{liu-etal-2019-multi}
Xiaodong Liu, Pengcheng He, Weizhu Chen, and Jianfeng Gao. 2019{\natexlab{a}}.
\newblock \href {https://doi.org/10.18653/v1/P19-1441} {Multi-task deep neural
  networks for natural language understanding}.
\newblock In \emph{Proceedings of the 57th Annual Meeting of the Association
  for Computational Linguistics}, pages 4487--4496, Florence, Italy.
  Association for Computational Linguistics.

\bibitem[{Liu et~al.(2019{\natexlab{b}})Liu, Ott, Goyal, Du, Joshi, Chen, Levy,
  Lewis, Zettlemoyer, and Stoyanov}]{DBLP:journals/corr/abs-1907-11692}
Yinhan Liu, Myle Ott, Naman Goyal, Jingfei Du, Mandar Joshi, Danqi Chen, Omer
  Levy, Mike Lewis, Luke Zettlemoyer, and Veselin Stoyanov. 2019{\natexlab{b}}.
\newblock \href {http://arxiv.org/abs/1907.11692} {Roberta: {A} robustly
  optimized {BERT} pretraining approach}.
\newblock \emph{CoRR}, abs/1907.11692.

\bibitem[{Mahabadi et~al.(2021)Mahabadi, Henderson, and
  Ruder}]{DBLP:conf/nips/MahabadiHR21}
Rabeeh~Karimi Mahabadi, James Henderson, and Sebastian Ruder. 2021.
\newblock \href
  {https://proceedings.neurips.cc/paper/2021/hash/081be9fdff07f3bc808f935906ef70c0-Abstract.html}
  {Compacter: Efficient low-rank hypercomplex adapter layers}.
\newblock In \emph{Advances in Neural Information Processing Systems 34: Annual
  Conference on Neural Information Processing Systems 2021, NeurIPS 2021,
  December 6-14, 2021, virtual}, pages 1022--1035.

\bibitem[{McCloskey and Cohen(1989)}]{MCCLOSKEY1989109}
Michael McCloskey and Neal~J. Cohen. 1989.
\newblock \href {https://doi.org/https://doi.org/10.1016/S0079-7421(08)60536-8}
  {Catastrophic interference in connectionist networks: The sequential learning
  problem}.
\newblock volume~24 of \emph{Psychology of Learning and Motivation}, pages
  109--165. Academic Press.

\bibitem[{Mhaske et~al.(2022)Mhaske, Kedia, Doddapaneni, Khapra, Kumar, Murthy,
  and Kunchukuttan}]{DBLP:journals/corr/abs-2212-10168}
Arnav Mhaske, Harshit Kedia, Sumanth Doddapaneni, Mitesh~M. Khapra, Pratyush
  Kumar, V.~Rudra Murthy, and Anoop Kunchukuttan. 2022.
\newblock \href {https://doi.org/10.48550/arXiv.2212.10168} {Naamapadam: {A}
  large-scale named entity annotated data for indic languages}.
\newblock \emph{CoRR}, abs/2212.10168.

\bibitem[{Muennighoff et~al.(2022)Muennighoff, Wang, Sutawika, Roberts,
  Biderman, Scao, Bari, Shen, Yong, Schoelkopf, Tang, Radev, Aji, Almubarak,
  Albanie, Alyafeai, Webson, Raff, and Raffel}]{muennighoff2022crosslingual}
Niklas Muennighoff, Thomas Wang, Lintang Sutawika, Adam Roberts, Stella
  Biderman, Teven~Le Scao, M~Saiful Bari, Sheng Shen, Zheng-Xin Yong, Hailey
  Schoelkopf, Xiangru Tang, Dragomir Radev, Alham~Fikri Aji, Khalid Almubarak,
  Samuel Albanie, Zaid Alyafeai, Albert Webson, Edward Raff, and Colin Raffel.
  2022.
\newblock \href {http://arxiv.org/abs/2211.01786} {Crosslingual generalization
  through multitask finetuning}.

\bibitem[{Ogueji et~al.(2021)Ogueji, Zhu, and Lin}]{ogueji-etal-2021-small}
Kelechi Ogueji, Yuxin Zhu, and Jimmy Lin. 2021.
\newblock \href {https://doi.org/10.18653/v1/2021.mrl-1.11} {Small data? no
  problem! exploring the viability of pretrained multilingual language models
  for low-resourced languages}.
\newblock In \emph{Proceedings of the 1st Workshop on Multilingual
  Representation Learning}, pages 116--126, Punta Cana, Dominican Republic.
  Association for Computational Linguistics.

\bibitem[{Pfeiffer et~al.(2021)Pfeiffer, Kamath, R{\"u}ckl{\'e}, Cho, and
  Gurevych}]{pfeiffer-etal-2021-adapterfusion}
Jonas Pfeiffer, Aishwarya Kamath, Andreas R{\"u}ckl{\'e}, Kyunghyun Cho, and
  Iryna Gurevych. 2021.
\newblock \href {https://doi.org/10.18653/v1/2021.eacl-main.39}
  {{A}dapter{F}usion: Non-destructive task composition for transfer learning}.
\newblock In \emph{Proceedings of the 16th Conference of the European Chapter
  of the Association for Computational Linguistics: Main Volume}, pages
  487--503, Online. Association for Computational Linguistics.

\bibitem[{Pfeiffer et~al.(2020{\natexlab{a}})Pfeiffer, R{\"u}ckl{\'e}, Poth,
  Kamath, Vuli{\'c}, Ruder, Cho, and Gurevych}]{pfeiffer-etal-2020-adapterhub}
Jonas Pfeiffer, Andreas R{\"u}ckl{\'e}, Clifton Poth, Aishwarya Kamath, Ivan
  Vuli{\'c}, Sebastian Ruder, Kyunghyun Cho, and Iryna Gurevych.
  2020{\natexlab{a}}.
\newblock \href {https://doi.org/10.18653/v1/2020.emnlp-demos.7}
  {{A}dapter{H}ub: A framework for adapting transformers}.
\newblock In \emph{Proceedings of the 2020 Conference on Empirical Methods in
  Natural Language Processing: System Demonstrations}, pages 46--54, Online.
  Association for Computational Linguistics.

\bibitem[{Pfeiffer et~al.(2020{\natexlab{b}})Pfeiffer, Vuli{\'c}, Gurevych, and
  Ruder}]{pfeiffer-etal-2020-mad}
Jonas Pfeiffer, Ivan Vuli{\'c}, Iryna Gurevych, and Sebastian Ruder.
  2020{\natexlab{b}}.
\newblock \href {https://doi.org/10.18653/v1/2020.emnlp-main.617} {{MAD-X}:
  {A}n {A}dapter-{B}ased {F}ramework for {M}ulti-{T}ask {C}ross-{L}ingual
  {T}ransfer}.
\newblock In \emph{Proceedings of the 2020 Conference on Empirical Methods in
  Natural Language Processing (EMNLP)}, pages 7654--7673, Online. Association
  for Computational Linguistics.

\bibitem[{Rajpurkar et~al.(2016)Rajpurkar, Zhang, Lopyrev, and
  Liang}]{rajpurkar-etal-2016-squad}
Pranav Rajpurkar, Jian Zhang, Konstantin Lopyrev, and Percy Liang. 2016.
\newblock \href {https://doi.org/10.18653/v1/D16-1264} {{SQ}u{AD}: 100,000+
  questions for machine comprehension of text}.
\newblock In \emph{Proceedings of the 2016 Conference on Empirical Methods in
  Natural Language Processing}, pages 2383--2392, Austin, Texas. Association
  for Computational Linguistics.

\bibitem[{R{\"u}ckl{\'e} et~al.(2021)R{\"u}ckl{\'e}, Geigle, Glockner, Beck,
  Pfeiffer, Reimers, and Gurevych}]{ruckle-etal-2021-adapterdrop}
Andreas R{\"u}ckl{\'e}, Gregor Geigle, Max Glockner, Tilman Beck, Jonas
  Pfeiffer, Nils Reimers, and Iryna Gurevych. 2021.
\newblock \href {https://doi.org/10.18653/v1/2021.emnlp-main.626}
  {{AdapterDrop}: {O}n the efficiency of adapters in transformers}.
\newblock In \emph{Proceedings of the 2021 Conference on Empirical Methods in
  Natural Language Processing}, pages 7930--7946, Online and Punta Cana,
  Dominican Republic. Association for Computational Linguistics.

\bibitem[{Ruder(2017)}]{DBLP:journals/corr/Ruder17a}
Sebastian Ruder. 2017.
\newblock \href {http://arxiv.org/abs/1706.05098} {An overview of multi-task
  learning in deep neural networks}.
\newblock \emph{CoRR}, abs/1706.05098.

\bibitem[{Sanh et~al.(2022)Sanh, Webson, Raffel, Bach, Sutawika, Alyafeai,
  Chaffin, Stiegler, Raja, Dey, Bari, Xu, Thakker, Sharma, Szczechla, Kim,
  Chhablani, Nayak, Datta, Chang, Jiang, Wang, Manica, Shen, Yong, Pandey,
  Bawden, Wang, Neeraj, Rozen, Sharma, Santilli, Fevry, Fries, Teehan, Scao,
  Biderman, Gao, Wolf, and Rush}]{sanh2022multitask}
Victor Sanh, Albert Webson, Colin Raffel, Stephen Bach, Lintang Sutawika, Zaid
  Alyafeai, Antoine Chaffin, Arnaud Stiegler, Arun Raja, Manan Dey, M~Saiful
  Bari, Canwen Xu, Urmish Thakker, Shanya~Sharma Sharma, Eliza Szczechla,
  Taewoon Kim, Gunjan Chhablani, Nihal Nayak, Debajyoti Datta, Jonathan Chang,
  Mike Tian-Jian Jiang, Han Wang, Matteo Manica, Sheng Shen, Zheng~Xin Yong,
  Harshit Pandey, Rachel Bawden, Thomas Wang, Trishala Neeraj, Jos Rozen,
  Abheesht Sharma, Andrea Santilli, Thibault Fevry, Jason~Alan Fries, Ryan
  Teehan, Teven~Le Scao, Stella Biderman, Leo Gao, Thomas Wolf, and Alexander~M
  Rush. 2022.
\newblock \href {https://openreview.net/forum?id=9Vrb9D0WI4} {Multitask
  prompted training enables zero-shot task generalization}.
\newblock In \emph{International Conference on Learning Representations}.

\bibitem[{Sap et~al.(2019)Sap, Rashkin, Chen, Le~Bras, and
  Choi}]{sap-etal-2019-social}
Maarten Sap, Hannah Rashkin, Derek Chen, Ronan Le~Bras, and Yejin Choi. 2019.
\newblock \href {https://doi.org/10.18653/v1/D19-1454} {Social {IQ}a:
  Commonsense reasoning about social interactions}.
\newblock In \emph{Proceedings of the 2019 Conference on Empirical Methods in
  Natural Language Processing and the 9th International Joint Conference on
  Natural Language Processing (EMNLP-IJCNLP)}, pages 4463--4473, Hong Kong,
  China. Association for Computational Linguistics.

\bibitem[{Srivastava et~al.(2022)Srivastava, Rastogi, Rao, Shoeb, Abid, Fisch,
  Brown, Santoro, Gupta, Garriga-Alonso, Kluska, Lewkowycz, Agarwal, Power,
  Ray, Warstadt, Kocurek, Safaya, Tazarv, Xiang, Parrish, Nie, Hussain, Askell,
  Dsouza, Slone, Rahane, Iyer, Andreassen, Madotto, Santilli, Stuhlmüller,
  Dai, La, Lampinen, Zou, Jiang, Chen, Vuong, Gupta, Gottardi, Norelli,
  Venkatesh, Gholamidavoodi, Tabassum, Menezes, Kirubarajan, Mullokandov,
  Sabharwal, Herrick, Efrat, Erdem, Karakaş, Roberts, Loe, Zoph, Bojanowski,
  Özyurt, Hedayatnia, Neyshabur, Inden, Stein, Ekmekci, Lin, Howald, Diao,
  Dour, Stinson, Argueta, Ramírez, Singh, Rathkopf, Meng, Baral, Wu,
  Callison-Burch, Waites, Voigt, Manning, Potts, Ramirez, Rivera, Siro, Raffel,
  Ashcraft, Garbacea, Sileo, Garrette, Hendrycks, Kilman, Roth, Freeman,
  Khashabi, Levy, González, Perszyk, Hernandez, Chen, Ippolito, Gilboa, Dohan,
  Drakard, Jurgens, Datta, Ganguli, Emelin, Kleyko, Yuret, Chen, Tam, Hupkes,
  Misra, Buzan, Mollo, Yang, Lee, Shutova, Cubuk, Segal, Hagerman, Barnes,
  Donoway, Pavlick, Rodola, Lam, Chu, Tang, Erdem, Chang, Chi, Dyer, Jerzak,
  Kim, Manyasi, Zheltonozhskii, Xia, Siar, Martínez-Plumed, Happé, Chollet,
  Rong, Mishra, Winata, de~Melo, Kruszewski, Parascandolo, Mariani, Wang,
  Jaimovitch-López, Betz, Gur-Ari, Galijasevic, Kim, Rashkin, Hajishirzi,
  Mehta, Bogar, Shevlin, Schütze, Yakura, Zhang, Wong, Ng, Noble, Jumelet,
  Geissinger, Kernion, Hilton, Lee, Fisac, Simon, Koppel, Zheng, Zou, Kocoń,
  Thompson, Kaplan, Radom, Sohl-Dickstein, Phang, Wei, Yosinski, Novikova,
  Bosscher, Marsh, Kim, Taal, Engel, Alabi, Xu, Song, Tang, Waweru, Burden,
  Miller, Balis, Berant, Frohberg, Rozen, Hernandez-Orallo, Boudeman, Jones,
  Tenenbaum, Rule, Chua, Kanclerz, Livescu, Krauth, Gopalakrishnan, Ignatyeva,
  Markert, Dhole, Gimpel, Omondi, Mathewson, Chiafullo, Shkaruta, Shridhar,
  McDonell, Richardson, Reynolds, Gao, Zhang, Dugan, Qin, Contreras-Ochando,
  Morency, Moschella, Lam, Noble, Schmidt, He, Colón, Metz, Şenel, Bosma,
  Sap, ter Hoeve, Farooqi, Faruqui, Mazeika, Baturan, Marelli, Maru, Quintana,
  Tolkiehn, Giulianelli, Lewis, Potthast, Leavitt, Hagen, Schubert,
  Baitemirova, Arnaud, McElrath, Yee, Cohen, Gu, Ivanitskiy, Starritt, Strube,
  Swędrowski, Bevilacqua, Yasunaga, Kale, Cain, Xu, Suzgun, Tiwari, Bansal,
  Aminnaseri, Geva, Gheini, T, Peng, Chi, Lee, Krakover, Cameron, Roberts,
  Doiron, Nangia, Deckers, Muennighoff, Keskar, Iyer, Constant, Fiedel, Wen,
  Zhang, Agha, Elbaghdadi, Levy, Evans, Casares, Doshi, Fung, Liang, Vicol,
  Alipoormolabashi, Liao, Liang, Chang, Eckersley, Htut, Hwang, Miłkowski,
  Patil, Pezeshkpour, Oli, Mei, Lyu, Chen, Banjade, Rudolph, Gabriel, Habacker,
  Delgado, Millière, Garg, Barnes, Saurous, Arakawa, Raymaekers, Frank,
  Sikand, Novak, Sitelew, LeBras, Liu, Jacobs, Zhang, Salakhutdinov, Chi, Lee,
  Stovall, Teehan, Yang, Singh, Mohammad, Anand, Dillavou, Shleifer, Wiseman,
  Gruetter, Bowman, Schoenholz, Han, Kwatra, Rous, Ghazarian, Ghosh, Casey,
  Bischoff, Gehrmann, Schuster, Sadeghi, Hamdan, Zhou, Srivastava, Shi, Singh,
  Asaadi, Gu, Pachchigar, Toshniwal, Upadhyay, Shyamolima, Debnath, Shakeri,
  Thormeyer, Melzi, Reddy, Makini, Lee, Torene, Hatwar, Dehaene, Divic, Ermon,
  Biderman, Lin, Prasad, Piantadosi, Shieber, Misherghi, Kiritchenko, Mishra,
  Linzen, Schuster, Li, Yu, Ali, Hashimoto, Wu, Desbordes, Rothschild, Phan,
  Wang, Nkinyili, Schick, Kornev, Telleen-Lawton, Tunduny, Gerstenberg, Chang,
  Neeraj, Khot, Shultz, Shaham, Misra, Demberg, Nyamai, Raunak, Ramasesh,
  Prabhu, Padmakumar, Srikumar, Fedus, Saunders, Zhang, Vossen, Ren, Tong,
  Zhao, Wu, Shen, Yaghoobzadeh, Lakretz, Song, Bahri, Choi, Yang, Hao, Chen,
  Belinkov, Hou, Hou, Bai, Seid, Zhao, Wang, Wang, Wang, and
  Wu}]{srivastava2022imitation}
Aarohi Srivastava, Abhinav Rastogi, Abhishek Rao, Abu Awal~Md Shoeb, Abubakar
  Abid, Adam Fisch, Adam~R. Brown, Adam Santoro, Aditya Gupta, Adrià
  Garriga-Alonso, Agnieszka Kluska, Aitor Lewkowycz, Akshat Agarwal, Alethea
  Power, Alex Ray, Alex Warstadt, Alexander~W. Kocurek, Ali Safaya, Ali Tazarv,
  Alice Xiang, Alicia Parrish, Allen Nie, Aman Hussain, Amanda Askell, Amanda
  Dsouza, Ambrose Slone, Ameet Rahane, Anantharaman~S. Iyer, Anders Andreassen,
  Andrea Madotto, Andrea Santilli, Andreas Stuhlmüller, Andrew Dai, Andrew La,
  Andrew Lampinen, Andy Zou, Angela Jiang, Angelica Chen, Anh Vuong, Animesh
  Gupta, Anna Gottardi, Antonio Norelli, Anu Venkatesh, Arash Gholamidavoodi,
  Arfa Tabassum, Arul Menezes, Arun Kirubarajan, Asher Mullokandov, Ashish
  Sabharwal, Austin Herrick, Avia Efrat, Aykut Erdem, Ayla Karakaş, B.~Ryan
  Roberts, Bao~Sheng Loe, Barret Zoph, Bartłomiej Bojanowski, Batuhan Özyurt,
  Behnam Hedayatnia, Behnam Neyshabur, Benjamin Inden, Benno Stein, Berk
  Ekmekci, Bill~Yuchen Lin, Blake Howald, Cameron Diao, Cameron Dour, Catherine
  Stinson, Cedrick Argueta, César~Ferri Ramírez, Chandan Singh, Charles
  Rathkopf, Chenlin Meng, Chitta Baral, Chiyu Wu, Chris Callison-Burch, Chris
  Waites, Christian Voigt, Christopher~D. Manning, Christopher Potts, Cindy
  Ramirez, Clara~E. Rivera, Clemencia Siro, Colin Raffel, Courtney Ashcraft,
  Cristina Garbacea, Damien Sileo, Dan Garrette, Dan Hendrycks, Dan Kilman, Dan
  Roth, Daniel Freeman, Daniel Khashabi, Daniel Levy, Daniel~Moseguí
  González, Danielle Perszyk, Danny Hernandez, Danqi Chen, Daphne Ippolito,
  Dar Gilboa, David Dohan, David Drakard, David Jurgens, Debajyoti Datta, Deep
  Ganguli, Denis Emelin, Denis Kleyko, Deniz Yuret, Derek Chen, Derek Tam,
  Dieuwke Hupkes, Diganta Misra, Dilyar Buzan, Dimitri~Coelho Mollo, Diyi Yang,
  Dong-Ho Lee, Ekaterina Shutova, Ekin~Dogus Cubuk, Elad Segal, Eleanor
  Hagerman, Elizabeth Barnes, Elizabeth Donoway, Ellie Pavlick, Emanuele
  Rodola, Emma Lam, Eric Chu, Eric Tang, Erkut Erdem, Ernie Chang, Ethan~A.
  Chi, Ethan Dyer, Ethan Jerzak, Ethan Kim, Eunice~Engefu Manyasi, Evgenii
  Zheltonozhskii, Fanyue Xia, Fatemeh Siar, Fernando Martínez-Plumed,
  Francesca Happé, Francois Chollet, Frieda Rong, Gaurav Mishra, Genta~Indra
  Winata, Gerard de~Melo, Germán Kruszewski, Giambattista Parascandolo,
  Giorgio Mariani, Gloria Wang, Gonzalo Jaimovitch-López, Gregor Betz, Guy
  Gur-Ari, Hana Galijasevic, Hannah Kim, Hannah Rashkin, Hannaneh Hajishirzi,
  Harsh Mehta, Hayden Bogar, Henry Shevlin, Hinrich Schütze, Hiromu Yakura,
  Hongming Zhang, Hugh~Mee Wong, Ian Ng, Isaac Noble, Jaap Jumelet, Jack
  Geissinger, Jackson Kernion, Jacob Hilton, Jaehoon Lee, Jaime~Fernández
  Fisac, James~B. Simon, James Koppel, James Zheng, James Zou, Jan Kocoń, Jana
  Thompson, Jared Kaplan, Jarema Radom, Jascha Sohl-Dickstein, Jason Phang,
  Jason Wei, Jason Yosinski, Jekaterina Novikova, Jelle Bosscher, Jennifer
  Marsh, Jeremy Kim, Jeroen Taal, Jesse Engel, Jesujoba Alabi, Jiacheng Xu,
  Jiaming Song, Jillian Tang, Joan Waweru, John Burden, John Miller, John~U.
  Balis, Jonathan Berant, Jörg Frohberg, Jos Rozen, Jose Hernandez-Orallo,
  Joseph Boudeman, Joseph Jones, Joshua~B. Tenenbaum, Joshua~S. Rule, Joyce
  Chua, Kamil Kanclerz, Karen Livescu, Karl Krauth, Karthik Gopalakrishnan,
  Katerina Ignatyeva, Katja Markert, Kaustubh~D. Dhole, Kevin Gimpel, Kevin
  Omondi, Kory Mathewson, Kristen Chiafullo, Ksenia Shkaruta, Kumar Shridhar,
  Kyle McDonell, Kyle Richardson, Laria Reynolds, Leo Gao, Li~Zhang, Liam
  Dugan, Lianhui Qin, Lidia Contreras-Ochando, Louis-Philippe Morency, Luca
  Moschella, Lucas Lam, Lucy Noble, Ludwig Schmidt, Luheng He, Luis~Oliveros
  Colón, Luke Metz, Lütfi~Kerem Şenel, Maarten Bosma, Maarten Sap, Maartje
  ter Hoeve, Maheen Farooqi, Manaal Faruqui, Mantas Mazeika, Marco Baturan,
  Marco Marelli, Marco Maru, Maria Jose~Ramírez Quintana, Marie Tolkiehn,
  Mario Giulianelli, Martha Lewis, Martin Potthast, Matthew~L. Leavitt,
  Matthias Hagen, Mátyás Schubert, Medina~Orduna Baitemirova, Melody Arnaud,
  Melvin McElrath, Michael~A. Yee, Michael Cohen, Michael Gu, Michael
  Ivanitskiy, Michael Starritt, Michael Strube, Michał Swędrowski, Michele
  Bevilacqua, Michihiro Yasunaga, Mihir Kale, Mike Cain, Mimee Xu, Mirac
  Suzgun, Mo~Tiwari, Mohit Bansal, Moin Aminnaseri, Mor Geva, Mozhdeh Gheini,
  Mukund~Varma T, Nanyun Peng, Nathan Chi, Nayeon Lee, Neta Gur-Ari Krakover,
  Nicholas Cameron, Nicholas Roberts, Nick Doiron, Nikita Nangia, Niklas
  Deckers, Niklas Muennighoff, Nitish~Shirish Keskar, Niveditha~S. Iyer, Noah
  Constant, Noah Fiedel, Nuan Wen, Oliver Zhang, Omar Agha, Omar Elbaghdadi,
  Omer Levy, Owain Evans, Pablo Antonio~Moreno Casares, Parth Doshi, Pascale
  Fung, Paul~Pu Liang, Paul Vicol, Pegah Alipoormolabashi, Peiyuan Liao, Percy
  Liang, Peter Chang, Peter Eckersley, Phu~Mon Htut, Pinyu Hwang, Piotr
  Miłkowski, Piyush Patil, Pouya Pezeshkpour, Priti Oli, Qiaozhu Mei, Qing
  Lyu, Qinlang Chen, Rabin Banjade, Rachel~Etta Rudolph, Raefer Gabriel, Rahel
  Habacker, Ramón~Risco Delgado, Raphaël Millière, Rhythm Garg, Richard
  Barnes, Rif~A. Saurous, Riku Arakawa, Robbe Raymaekers, Robert Frank, Rohan
  Sikand, Roman Novak, Roman Sitelew, Ronan LeBras, Rosanne Liu, Rowan Jacobs,
  Rui Zhang, Ruslan Salakhutdinov, Ryan Chi, Ryan Lee, Ryan Stovall, Ryan
  Teehan, Rylan Yang, Sahib Singh, Saif~M. Mohammad, Sajant Anand, Sam
  Dillavou, Sam Shleifer, Sam Wiseman, Samuel Gruetter, Samuel~R. Bowman,
  Samuel~S. Schoenholz, Sanghyun Han, Sanjeev Kwatra, Sarah~A. Rous, Sarik
  Ghazarian, Sayan Ghosh, Sean Casey, Sebastian Bischoff, Sebastian Gehrmann,
  Sebastian Schuster, Sepideh Sadeghi, Shadi Hamdan, Sharon Zhou, Shashank
  Srivastava, Sherry Shi, Shikhar Singh, Shima Asaadi, Shixiang~Shane Gu, Shubh
  Pachchigar, Shubham Toshniwal, Shyam Upadhyay, Shyamolima, Debnath, Siamak
  Shakeri, Simon Thormeyer, Simone Melzi, Siva Reddy, Sneha~Priscilla Makini,
  Soo-Hwan Lee, Spencer Torene, Sriharsha Hatwar, Stanislas Dehaene, Stefan
  Divic, Stefano Ermon, Stella Biderman, Stephanie Lin, Stephen Prasad,
  Steven~T. Piantadosi, Stuart~M. Shieber, Summer Misherghi, Svetlana
  Kiritchenko, Swaroop Mishra, Tal Linzen, Tal Schuster, Tao Li, Tao Yu, Tariq
  Ali, Tatsu Hashimoto, Te-Lin Wu, Théo Desbordes, Theodore Rothschild, Thomas
  Phan, Tianle Wang, Tiberius Nkinyili, Timo Schick, Timofei Kornev, Timothy
  Telleen-Lawton, Titus Tunduny, Tobias Gerstenberg, Trenton Chang, Trishala
  Neeraj, Tushar Khot, Tyler Shultz, Uri Shaham, Vedant Misra, Vera Demberg,
  Victoria Nyamai, Vikas Raunak, Vinay Ramasesh, Vinay~Uday Prabhu, Vishakh
  Padmakumar, Vivek Srikumar, William Fedus, William Saunders, William Zhang,
  Wout Vossen, Xiang Ren, Xiaoyu Tong, Xinran Zhao, Xinyi Wu, Xudong Shen,
  Yadollah Yaghoobzadeh, Yair Lakretz, Yangqiu Song, Yasaman Bahri, Yejin Choi,
  Yichi Yang, Yiding Hao, Yifu Chen, Yonatan Belinkov, Yu~Hou, Yufang Hou,
  Yuntao Bai, Zachary Seid, Zhuoye Zhao, Zijian Wang, Zijie~J. Wang, Zirui
  Wang, and Ziyi Wu. 2022.
\newblock \href {http://arxiv.org/abs/2206.04615} {Beyond the imitation game:
  Quantifying and extrapolating the capabilities of language models}.

\bibitem[{Tjong Kim~Sang and
  De~Meulder(2003)}]{tjong-kim-sang-de-meulder-2003-introduction}
Erik~F. Tjong Kim~Sang and Fien De~Meulder. 2003.
\newblock \href {https://aclanthology.org/W03-0419} {Introduction to the
  {C}o{NLL}-2003 shared task: Language-independent named entity recognition}.
\newblock In \emph{Proceedings of the Seventh Conference on Natural Language
  Learning at {HLT}-{NAACL} 2003}, pages 142--147.

\bibitem[{Vaswani et~al.(2017)Vaswani, Shazeer, Parmar, Uszkoreit, Jones,
  Gomez, Kaiser, and Polosukhin}]{NIPS2017_3f5ee243}
Ashish Vaswani, Noam Shazeer, Niki Parmar, Jakob Uszkoreit, Llion Jones,
  Aidan~N Gomez, \L~ukasz Kaiser, and Illia Polosukhin. 2017.
\newblock \href
  {https://proceedings.neurips.cc/paper/2017/file/3f5ee243547dee91fbd053c1c4a845aa-Paper.pdf}
  {Attention is all you need}.
\newblock In \emph{Advances in Neural Information Processing Systems},
  volume~30. Curran Associates, Inc.

\bibitem[{Vu et~al.(2022)Vu, Barua, Lester, Cer, Iyyer, and
  Constant}]{DBLP:journals/corr/abs-2205-12647}
Tu~Vu, Aditya Barua, Brian Lester, Daniel Cer, Mohit Iyyer, and Noah Constant.
  2022.
\newblock \href {https://doi.org/10.48550/arXiv.2205.12647} {Overcoming
  catastrophic forgetting in zero-shot cross-lingual generation}.
\newblock \emph{CoRR}, abs/2205.12647.

\bibitem[{Wei et~al.(2021)Wei, Bosma, Zhao, Guu, Yu, Lester, Du, Dai, and
  Le}]{wei2021finetuned}
Jason Wei, Maarten Bosma, Vincent~Y. Zhao, Kelvin Guu, Adams~Wei Yu, Brian
  Lester, Nan Du, Andrew~M. Dai, and Quoc~V. Le. 2021.
\newblock \href {http://arxiv.org/abs/2109.01652} {Finetuned language models
  are zero-shot learners}.

\bibitem[{Williams et~al.(2018)Williams, Nangia, and
  Bowman}]{williams-etal-2018-broad}
Adina Williams, Nikita Nangia, and Samuel Bowman. 2018.
\newblock \href {https://doi.org/10.18653/v1/N18-1101} {A broad-coverage
  challenge corpus for sentence understanding through inference}.
\newblock In \emph{Proceedings of the 2018 Conference of the North {A}merican
  Chapter of the Association for Computational Linguistics: Human Language
  Technologies, Volume 1 (Long Papers)}, pages 1112--1122, New Orleans,
  Louisiana. Association for Computational Linguistics.

\bibitem[{Wolf et~al.(2020)Wolf, Debut, Sanh, Chaumond, Delangue, Moi, Cistac,
  Rault, Louf, Funtowicz, Davison, Shleifer, von Platen, Ma, Jernite, Plu, Xu,
  Le~Scao, Gugger, Drame, Lhoest, and Rush}]{wolf-etal-2020-transformers}
Thomas Wolf, Lysandre Debut, Victor Sanh, Julien Chaumond, Clement Delangue,
  Anthony Moi, Pierric Cistac, Tim Rault, Remi Louf, Morgan Funtowicz, Joe
  Davison, Sam Shleifer, Patrick von Platen, Clara Ma, Yacine Jernite, Julien
  Plu, Canwen Xu, Teven Le~Scao, Sylvain Gugger, Mariama Drame, Quentin Lhoest,
  and Alexander Rush. 2020.
\newblock \href {https://doi.org/10.18653/v1/2020.emnlp-demos.6} {Transformers:
  State-of-the-art natural language processing}.
\newblock In \emph{Proceedings of the 2020 Conference on Empirical Methods in
  Natural Language Processing: System Demonstrations}, pages 38--45, Online.
  Association for Computational Linguistics.

\bibitem[{Yang et~al.(2019)Yang, Zhang, Tar, and
  Baldridge}]{yang-etal-2019-paws}
Yinfei Yang, Yuan Zhang, Chris Tar, and Jason Baldridge. 2019.
\newblock \href {https://doi.org/10.18653/v1/D19-1382} {{PAWS}-{X}: A
  cross-lingual adversarial dataset for paraphrase identification}.
\newblock In \emph{Proceedings of the 2019 Conference on Empirical Methods in
  Natural Language Processing and the 9th International Joint Conference on
  Natural Language Processing (EMNLP-IJCNLP)}, pages 3687--3692, Hong Kong,
  China. Association for Computational Linguistics.

\bibitem[{Yong et~al.(2022)Yong, Schoelkopf, Muennighoff, Aji, Adelani,
  Almubarak, Bari, Sutawika, Kasai, Baruwa, Winata, Biderman, Radev, and
  Nikoulina}]{DBLP:journals/corr/abs-2212-09535}
Zheng~Xin Yong, Hailey Schoelkopf, Niklas Muennighoff, Alham~Fikri Aji,
  David~Ifeoluwa Adelani, Khalid Almubarak, M.~Saiful Bari, Lintang Sutawika,
  Jungo Kasai, Ahmed Baruwa, Genta~Indra Winata, Stella Biderman, Dragomir
  Radev, and Vassilina Nikoulina. 2022.
\newblock \href {https://doi.org/10.48550/arXiv.2212.09535} {{BLOOM+1:} adding
  language support to {BLOOM} for zero-shot prompting}.
\newblock \emph{CoRR}, abs/2212.09535.

\bibitem[{Zhang et~al.(2022)Zhang, Yu, Yu, Guo, and
  Jiang}]{DBLP:journals/corr/abs-2204-03508}
Zhihan Zhang, Wenhao Yu, Mengxia Yu, Zhichun Guo, and Meng Jiang. 2022.
\newblock \href {https://doi.org/10.48550/arXiv.2204.03508} {A survey of
  multi-task learning in natural language processing: Regarding task
  relatedness and training methods}.
\newblock \emph{CoRR}, abs/2204.03508.

\bibitem[{Zoph et~al.(2016)Zoph, Yuret, May, and
  Knight}]{zoph-etal-2016-transfer}
Barret Zoph, Deniz Yuret, Jonathan May, and Kevin Knight. 2016.
\newblock \href {https://doi.org/10.18653/v1/D16-1163} {Transfer learning for
  low-resource neural machine translation}.
\newblock In \emph{Proceedings of the 2016 Conference on Empirical Methods in
  Natural Language Processing}, pages 1568--1575, Austin, Texas. Association
  for Computational Linguistics.

\end{thebibliography}
\bibliographystyle{acl_natbib}
\section{Appendices} 
\label{sec:appendix}

\subsection{Details of Tasks and Languages}
\label{sec:appendix_tasklanguagesdetails}

\noindent \textbf{Sentence Classification} tasks are Natural Language Inference (NLI), sentiment classification, paraphrase detection and Choice Of Plausible Alternatives (COPA). 
For NLI we use the MultiNLI \cite{williams-etal-2018-broad} dataset for training and test performance on IndicXNLI for 11 languages. For sentiment classification, we train on the Amazon Multilingual Reviews (AMR) dataset \cite{DBLP:conf/emnlp/KeungLSS20} and test on IndicSentiment for 11 languages. For paraphrase detection, we train on the PAWS-X \cite{yang-etal-2019-paws} dataset and test on IndicXParaphrase  for 10 languages. For the COPA task, which involves selecting one of two alternatives that more plausibly has a causal relation with a given premise, we train on SocialIQA \cite{sap-etal-2019-social} and test on IndicCOPA for 11 languages. 

\noindent \textbf{Token Classification} task uses the CoNLL-2003  \cite{tjong-kim-sang-de-meulder-2003-introduction} dataset for training and Naamapadam \cite{DBLP:journals/corr/abs-2212-10168} for testing for 11 languages.

\noindent \textbf{Question Answering}
We use the SQuAD \cite{rajpurkar-etal-2016-squad} data for training and test on the IndicQA benchmark \cite{DBLP:journals/corr/abs-2212-05409} available in 11 Indian languages.

\subsection{Task-level sensitivity}
\label{sec:appendix_tasklevelsensitivity}
The efficiency of training is also affected by the task, as shown in Table~\ref{compute-percent: indicbert}, where the QA task requires relatively more FLOs compared to the paraphrase task. However, across all tasks the trend remains the same. 
\begin{table*}[t]
\small
\centering
\begin{tabular}{lccccccc}
\toprule
\textbf{Method} & \textbf{Sentiment} &\textbf{XNLI} & \textbf{COPA}& \textbf{Paraphrase} & \textbf{NER} & \textbf{QA} & \textbf{Total} \\
\midrule
Houlsby & 1.8E+17 & 4.0E+17 & 3.8E+17 & 1.5E+17 & 4.2E+15 & 7.3E+17 & 1.8E+18 \\
Bapna & 1.5E+17 & 4.5E+17 & 3.3E+17 & 1.1E+17 & 6.2E+15 & 5.8E+17 & 1.6E+18 \\
Houlsby Parallel & 1.1E+17 & 4.0E+17 & 3.0E+17 & 7.4E+16 & 6.6E+15 & 3.9E+17 & 1.3E+18 \\
Bapna Parallel & 8.6E+16 & 3.9E+17 & 2.3E+17 & 1.2E+17 & 4.4E+15 & 5.1E+17 & 1.3E+18 \\
Prefix Tuning & 1.5E+17 & 4.4E+17 & 2.2E+17 & 1.5E+17 & 6.2E+15 & 3.1E+17 & 1.3E+18 \\
Lora & 1.7E+17 & 3.9E+17 & 2.1E+17 & 1.5E+17 & 8.5E+15 & 5.2E+17 & 1.5E+18 \\
Compacter & 2.4E+17 & 2.9E+17 & 5.9E+17 & 9.8E+16 & 1.2E+16 & 8.7E+17 & 2.1E+18 \\
Adapter Drop & 1.2E+17 & 4.2E+17 & 1.9E+17 & 4.6E+16 & 4.2E+15 & 1.1E+17 & 8.8E+17 \\
FT & 5.3E+16 & 1.3E+17 & 7.9E+16 & 7.8E+16 & 3.5E+15 & 1.0E+17 & 4.5E+17 \\ 
\midrule
Total  & 1.3E+18 & 3.3E+18 & 2.5E+18 & 9.8E+17 & 5.6E+16 & 4.1E+18 & 1.2E+19 \\
\bottomrule
\end{tabular}
\caption{The table reports the total FLOS for FT and various adapters on IndicBERT, across each of the tasks. Total corresponds to the total FLOS summed over all the tasks for a particular fine-tuning method.}
\label{compute-avg-absolute: indicbert}
\end{table*}



\begingroup
\setlength{\tabcolsep}{3pt} 
\renewcommand{\arraystretch}{1} 
\begin{table*}[]
\small
\centering
\begin{tabular}{llcccccccc}
\toprule
\begin{tabular}[c]{@{}c@{}}\textbf{Target}\\ \textbf{Task}\end{tabular} & \textbf{Step} & \begin{tabular}[c]{@{}c@{}}\textbf{Amazon Multi}\\ \textbf{Reviews}\end{tabular} & \textbf{XNLI} & \textbf{COPA} & \textbf{PAWS} & \textbf{CoNLL2003} & \textbf{SQuAD} & \textbf{Avg -1} & \textbf{Avg} \\

\midrule
Baseline & Full FT & 93.8 & \textbf{83.0} & 62.3 & 93.0 & \textbf{92.8} & 82.1 & - & \textbf{84.5} \\
 & MTL (full) & 93.5 & 80.9 & 61.4 & 91.5 & 91.0 & 82.1 & - & 83.4 \\
 \midrule
 Best Adapter & Houlsby & 94.0 & 82.4 & 61.5 & 92.3 & 91.5 & 81.7 & -&  83.9  \\
 \midrule
Sentiment & MTL$_{-1}$ & - & 81.6 & \cellcolor{blue!25}\textbf{63.0} & 91.5 & \cellcolor{blue!25}92.5 & \cellcolor{blue!25}82.5 & 82.2 & - \\
 &{MTL}$_{+tgt+old}$ & 93.1 & 79.0 & 60.7 & 89.0 & 91.4 & 81.3 & 80.3 & 82.4 \\
 & MTL$_{+tgt}$ & 93.5 & 58.6 & 47.1 & 71.4 & 78.3 & 71.7 & 65.4 & 70.1 \\
 \midrule
XNLI & MTL$_{-1}$ & \cellcolor{blue!25}\textbf{94.1} & - & 60.5 & 91.5 & \cellcolor{blue!25}91.9 & \cellcolor{blue!25}82.4 & 84.1 & - \\
 & {MTL}$_{+tgt+old}$ & 92.9 & 79.0 & 58.7 & 88.3 & 87.7 & 78.6 & 81.2 & 80.9 \\
 & MTL$_{+tgt}$ & 90.7 & 79.6 & 52.8 & 56.5 & 85.4 & 36.4 & 64.4 & 66.9 \\
\midrule
COPA &MTL$_{-1}$ & 93.8 & 81.9 & - & 91.6 & 91.0 & 81.5 & 88.0 & - \\
 & {MTL}$_{+tgt+old}$ & 93.5 & 79.0 & \cellcolor{blue!25}62.5 & 90.8 & 90.9 & 78.7 & 86.6 & 82.6 \\
 & MTL$_{+tgt}$ & 92.4 & 73.8 & \cellcolor{blue!25}62.2 & 87.8 & 89.8 & 79.2 & 84.6 & 80.9 \\
\midrule
Paraphrase & MTL$_{-1}$ & 93.9 & 80.1 & \cellcolor{blue!25}62.4 & - & \cellcolor{blue!25}91.9 & 80.9 & 81.8 & - \\
 & {MTL}$_{+tgt+old}$ & 93.9 & 79.8 & 60.2 & 89.7 & \cellcolor{blue!25}91.7 & 80.9 & 81.3 & 82.7 \\
 & MTL$_{+tgt}$ & 92.9 & 73.9 & 59.8 & 92.2 & 77.4 & 73.8 & 75.5 & 78.3 \\
\midrule
NER &  MTL$_{-1}$ & 94.0 & 82.1 & \cellcolor{blue!25}62.2 & 92.7 & - & \cellcolor{blue!25}\textbf{82.6} & 82.7 & - \\
 & {MTL}$_{+tgt+old}$ & 93.4 & 80.8 & 61.0 & 91.2 & 91.4 & 81.0 & 81.5 & 83.1 \\
 & MTL$_{+tgt}$ & 93.1 & 74.1 & 59.9 & 74.5 & \cellcolor{blue!25}92.1 & 71.6 & 74.6 & 77.6 \\
\midrule
QA & MTL$_{-1}$ & \cellcolor{blue!25}\textbf{94.1} & 81.6 & \cellcolor{blue!25}62.9 & \cellcolor{blue!25}\textbf{93.4} & \cellcolor{blue!25}92.0 & - & 84.8 & - \\
 & {MTL}$_{+tgt+old}$ & 93.5 & 80.0 & 59.7 & 91.4 & 89.9 & 81.2 & 82.9 & 82.6 \\
 & MTL$_{+tgt}$ & 92.4 & 77.5 & 61.0 & 73.0 & 63.5 & \cellcolor{blue!25}82.5 & 73.5 & 75.0 \\
 \bottomrule
\end{tabular}
\caption{This table reports \textbf{in-language (train and test on English)} performance for maintainability of MTL. "Target task" is held out task i.e. pre-trained IndicBERT model is fine-tuned on the remaining 5 task representing  MTL$_{-1}$ model. {MTL}$_{+tgt+old}$ represents continual fine-tuning of the MTL$_{-1}$ model on the target task dataset and 10\% of the existing task dataset. {MTL}$_{+tgt}$ represents continual fine-tuning of the MTL$_{-1}$ model on the target task dataset. "Avg -1 " reports the in-language performance averaged over the task included in MTL$_{-1}$ step. "Avg" reports the in-language performance averaged over all 6 task. Here, text bold indicates the best value in the column and colored cell represent MTL is performing better than the Best Adapter method.}
\label{table:MTL_en}
\end{table*}
\endgroup

\begingroup
\setlength{\tabcolsep}{2pt} 
\renewcommand{\arraystretch}{1} 
\begin{table*}[]
\small
\centering
\begin{tabular}{llccccccccc}
\toprule
\begin{tabular}[c]{@{}c@{}}\textbf{Target}\\ \textbf{Task}\end{tabular} & \textbf{Step} & \begin{tabular}[c]{@{}c@{}}\textbf{Indic}\\ \textbf{Sentiment}\end{tabular} & \begin{tabular}[c]{@{}c@{}}\textbf{Indic}\\ \textbf{XNLI}\end{tabular} & \begin{tabular}[c]{@{}c@{}}\textbf{Indic}\\ \textbf{COPA}\end{tabular} & \begin{tabular}[c]{@{}c@{}}\textbf{Indic}\\ \textbf{XPara}\end{tabular} & \begin{tabular}[c]{@{}c@{}}\textbf{Naama-}\\ \textbf{padam}\end{tabular} & \textbf{IndicQA} & \textbf{Avg -1} & \textbf{Avg} &  \begin{tabular}[c]{@{}c@{}}\textbf{\% $\uparrow$} \\\textbf{FLOs}\end{tabular} \\

\midrule
 Baseline & Full FT & \textbf{90.9} & \textbf{72.9} & 62.5 & 57.3 & 66.7 & 49.3 & - & 66.6&- \\
 & MTL & 88.5 & 71.2 & 64.9 & 74.0 & 65.8 & 45.4 & - & 68.3 &20.2 \\
 \midrule
 Best Adapter & Houlsby & 89.7 & \textbf{72.9} & 64.1 & 57.4 & 65.5 & \textbf{50.0} &-& 66.6 &311.7 \\
 \midrule
 
Sentiment & MTL$_{-1}$ & - & 71.5 & \cellcolor{blue!25}64.8 & \cellcolor{blue!25}74.8 & 65.1 & 46.9 & 64.6 & - \\
 & {MTL}$_{+tgt+old_{10}}$ & \cellcolor{blue!25}90.2 & 70.8 & 61.9 & \cellcolor{blue!25}72.9 & \cellcolor{blue!25}66.1 & 48.8 & 64.1 & \cellcolor{blue!25}\textbf{68.4} &2.3 \\
 & {MTL}$_{+tgt+old_{5}}$ & \cellcolor{blue!25}90.5 & 69.1 & 62.3 & \cellcolor{blue!25}74.2 & 62.2 & 47.3 & 63.0 & \cellcolor{blue!25}67.6 &  -0.4 \\
 & MTL$_{+tgt}$ & 89.1 & 54.9 & 52.2 & \cellcolor{blue!25}67.3 & 40.4 & 34.9 & 50.0 & 56.5 & 1.7 \\
 \midrule
 
XNLI & MTL$_{-1}$ & \cellcolor{blue!25}90.5 & - & \cellcolor{blue!25}\textbf{67.8} & 56.7 & 63.6 & 47.4 & 65.2 & - \\
 & {MTL}$_{+tgt+old_{10}}$ & \cellcolor{blue!25}90.8 & 71.2 & 63.6 & \cellcolor{blue!25}68.7 & 59.6 & 48.3 & 66.2 & \cellcolor{blue!25}67.0 & 20.5\\
 & {MTL}$_{+tgt+old_{5}}$ & \cellcolor{blue!25}90.5 & 70.6 & \cellcolor{blue!25}64.7 & \cellcolor{blue!25}61.9 & 63.5 & 47.3 & 65.6 & 66.4 & -2.6\\
 & MTL$_{+tgt}$ & 86.0 & 70.5 & \cellcolor{blue!25}64.5 & \cellcolor{blue!25}73.3 & 56.2 & 15.1 & 59.0 & 60.9 & 12.1\\
 \midrule
 
COPA & MTL$_{-1}$ & 88.8 & 72.3 & - & \cellcolor{blue!25}73.7 & 65.0 & 48.4 & 69.7 & - \\
 & {MTL}$_{+tgt+old_{10}}$ & 88.3 & 69.7 & \cellcolor{blue!25}65.6 & \cellcolor{blue!25}74.8 & 65.4 & 43.9 & 68.4 & \cellcolor{blue!25}67.9 & 15.3 \\
 & {MTL}$_{+tgt+old_{5}}$ & \cellcolor{blue!25}90.5 & 71.1 & \cellcolor{blue!25}64.6 & \cellcolor{blue!25}73.0 & 63.7 & 44.9 & 68.6 & \cellcolor{blue!25}68.0 & 3.5 \\
 & {MTL}$_{+tgt+old+min_{10}}$ & 85.5 & 69.7 & \cellcolor{blue!25}66.5 & \cellcolor{blue!25}74.6 & 63.6 & 45.7 & 67.8 & \cellcolor{blue!25}67.6 & 10.4 \\
 & {MTL}$_{+tgt+old+min_{5}}$ & \cellcolor{blue!25}90.0 & 69.8 & \cellcolor{blue!25}65.7 & \cellcolor{blue!25}73.9 & 63.8 & 46.2 & 68.7 & \cellcolor{blue!25}68.2 & 10.8\\
 & MTL$_{+tgt}$ & 89.5 & 66.4 & \cellcolor{blue!25}66.0 & \cellcolor{blue!25}75.5 & 62.7 & 46.4 & 68.1 & \cellcolor{blue!25}67.7 & 9.3\\
 \midrule
 
Paraphrase & MTL$_{-1}$ & 86.0 & 70.2 & \cellcolor{blue!25}64.4 & - & 65.0 & 45.0 & 66.1 & - \\
 & {MTL}$_{+tgt+old_{10}}$ & 87.4 & 69.8 & \cellcolor{blue!25}64.2 & \cellcolor{blue!25}77.8 & 65.0 & 45.4 & 66.4 & \cellcolor{blue!25}68.3 & 32.4 \\
 & {MTL}$_{+tgt+old_{5}}$ & 86.2 & 70.0 & \cellcolor{blue!25}64.8 & \cellcolor{blue!25}\textbf{78.0} & 64.2 & 42.7 & 65.6 & \cellcolor{blue!25}67.7 & 25.3 \\
 & MTL$_{+tgt}$ & 81.1 & 66.0 & \cellcolor{blue!25}64.4 & \cellcolor{blue!25}73.1 & 30.1 & 42.5 & 56.8 & 59.5 & 24.1 \\
 \midrule
 
NER & MTL$_{-1}$ & 88.0 & 72.5 & \cellcolor{blue!25}65.7 & \cellcolor{blue!25}77.3 & - & 47.7 & 70.3 & - \\
 & {MTL}$_{+tgt+old_{10}}$ & 86.7 & 71.2 & \cellcolor{blue!25}64.4 & \cellcolor{blue!25}76.3 & 65.2 & 45.1 & 68.7 & \cellcolor{blue!25}68.1 & 68.2\\
 & {MTL}$_{+tgt+old_{5}}$ & 88.4 & 70.4 & 63.9 & \cellcolor{blue!25}73.2 & \cellcolor{blue!25}65.7 & 45.7 & 68.3 & \cellcolor{blue!25}67.9 & 66.6\\
 & {MTL}$_{+tgt+old+min_{10}}$ & 87.6 & 71.2 & \cellcolor{blue!25}64.7 & \cellcolor{blue!25}73.5 & \cellcolor{blue!25}66.2 & 43.7 & 68.1 & \cellcolor{blue!25}67.8 & 66.6\\
 & {MTL}$_{+tgt+old+min_{5}}$ & 87.8 & 71.2 & \cellcolor{blue!25}65.4 & \cellcolor{blue!25}71.3 & 65.4 & 45.8 & 68.3 & \cellcolor{blue!25}67.8 & 63.1\\
 & MTL$_{+tgt}$ & 83.8 & 67.9 & 62.3 & 57.3 & \cellcolor{blue!25}\textbf{68.5} & 39.8 & 62.2 & 63.2 & 59.8 \\
 \midrule
 
QA & MTL$_{-1}$ & 89.2 & 72.3 & \cellcolor{blue!25}64.9 & \cellcolor{blue!25}74.9 & 65.4 & - & 73.3 & - \\
 & {MTL}$_{+tgt+old_{10}}$ & 85.9 & 71.1 & 63.9 & \cellcolor{blue!25}75.7 & 62.3 & 46.8 & 71.8 & \cellcolor{blue!25}67.6 & 25.8\\
 & {MTL}$_{+tgt+old_{5}}$ & 87.9 & 71.6 & \cellcolor{blue!25}64.4 & \cellcolor{blue!25}75.1 & \cellcolor{blue!25}65.6 & 45.0 & 72.9 & \cellcolor{blue!25}68.3 & 22.1\\
 & MTL$_{+tgt}$ & 84.9 & 68.2 & \cellcolor{blue!25}65.9 & \cellcolor{blue!25}66.9 & 23.7 & 46.6 & 61.9 & 59.4 & 21.2 \\
 \bottomrule
\end{tabular}
\caption{Table reports \textbf{cross-lingual performance (train on English test on Indic)}. Row {MTL}$_{+tgt+old_{10}}$ and {MTL}$_{+tgt+old_{5}}$ denotes adding 10\% and 5\% of existing task data combine with new task dataset respectively. {MTL}$_{+tgt+old+min_{10}}$ denotes combining the existing task dataset size minimum(10\% data , target task dataset size) i.e. to ensure the existing task dataset is less or equal to new task dataset when combined. similarly {MTL}$_{+tgt+old+min_{5}}$ denote combining the existing task dataset size as minimum(5\% data , target task dataset size).  "Avg -1 " reports the cross-lingual perfromance averaged over the task included in MTL$_{-1}$ step. "Avg" reports the cross-lingual performance averaged over all 6 task. Here, column "\%$\uparrow$FLOs" reports the relative percent increase in the total computation cost for adding all 6 task with respect to the total computation cost of fine-tuning. Note, we have {MTL}$_{+tgt+old+min_{10}}$ and {MTL}$_{+tgt+old+min_{5}}$ only for NER and COPA dataset, as dataset size for NER and COPA is less. Here, text bold indicates the best value in the column and colored cell represent MTL is performing better than the Best Adapter method.}
\label{table:MTL_XL_new}
\end{table*}
\endgroup

\begingroup
\setlength{\tabcolsep}{2pt} 
\renewcommand{\arraystretch}{1} 
\begin{table*}[]
\small
\centering
\begin{tabular}{llcccccccc}
\toprule
\begin{tabular}[c]{@{}c@{}}\textbf{Target}\\ \textbf{Task}\end{tabular} & \textbf{Step} & \begin{tabular}[c]{@{}c@{}}\textbf{Amazon Multi}\\ \textbf{Reviews}\end{tabular} & \textbf{XNLI} & \textbf{COPA} & \textbf{PAWS} & \textbf{CoNLL2003} & \textbf{SQuAD} & \textbf{Avg -1} & \textbf{Avg} \\

\midrule
Baseline & Full FT & 93.8 & \textbf{83.0} & 62.3 & 93.0 & \textbf{92.8} & 82.1 & - & \textbf{84.5} \\
 & MTL (full) & 93.5 & 80.9 & 61.4 & 91.5 & 91.0 & 82.1 & - & 83.4 \\
 \midrule
 Best Adapter & Houlsby & 94.0 & 82.4 & 61.5 & 92.3 & 91.5 & 81.7 & -&  83.9  \\
 \midrule
Sentiment & MTL$_{-1}$ & - & 81.6 & \cellcolor{blue!25}63.0 & 91.5 & \cellcolor{blue!25}92.5 & \cellcolor{blue!25}82.5 & 82.2 & - \\
 & {MTL}$_{+tgt+old_{10}}$ & 93.1 & 79.0 & 60.7 & 89.0 & 91.4 & 81.3 & 80.3 & 82.4 \\
 & {MTL}$_{+tgt+old_{5}}$ & 93.0 & 78.7 & 60.2 & 91.0 & \cellcolor{blue!25}91.7 & 81.0 & 80.5 & 82.6 \\
 & MTL$_{+tgt}$ & 93.5 & 58.6 & 47.1 & 71.4 & 78.3 & 71.7 & 65.4 & 70.1 \\
 \midrule
XNLI & MTL$_{-1}$  & \cellcolor{blue!25}\textbf{94.1} & - & 60.5 & 91.5 & \cellcolor{blue!25}91.9 & \cellcolor{blue!25}82.4 & 84.1 & - \\
 & {MTL}$_{+tgt+old_{10}}$ & 92.9 & 79.0 & 58.7 & 88.3 & 87.7 & 78.6 & 81.2 & 80.9 \\
 & {MTL}$_{+tgt+old_{5}}$ & 93.1 & 77.1 & 59.0 & 86.1 & 89.1 & 79.9 & 81.4 & 80.7 \\
 & MTL$_{+tgt}$ & 90.7 & 79.6 & 52.8 & 56.5 & 85.4 & 36.4 & 64.4 & 66.9 \\
 \midrule
COPA &MTL$_{-1}$ & 93.8 & 81.9 & - & 91.6 & 91.0 & 81.5 & 88.0 & - \\
 & {MTL}$_{+tgt+old_{10}}$ & 93.5 & 79.0 & \cellcolor{blue!25}62.5 & 90.8 & 90.9 & 78.7 & 86.6 & 82.6 \\
 & {MTL}$_{+tgt+old_{5}}$ & 93.7 & 80.8 & 58.6 & 90.8 & \cellcolor{blue!25}91.7 & 81.1 & 87.6 & 82.8 \\
 & {MTL}$_{+tgt+old+min_{10}}$ & 93.2 & 79.3 & \cellcolor{blue!25}62.2 & 91.9 & 91.0 & 81.1 & 87.3 & 83.1 \\
 & {MTL}$_{+tgt+old+min_{5}}$ & 93.8 & 79.6 & \cellcolor{blue!25}\textbf{63.2} & 90.8 & 90.9 & 80.6 & 87.1 & 83.1 \\
 & MTL$_{+tgt}$ & 92.4 & 73.8 & 62.2 & 87.8 & 89.8 & 79.2 & 84.6 & 80.9 \\
 \midrule
Paraphrase & MTL$_{-1}$  & 93.9 & 80.1 & \cellcolor{blue!25}62.4 & - & \cellcolor{blue!25}91.9 & 80.9 & 81.8 & - \\
 & {MTL}$_{+tgt+old_{10}}$ & 93.9 & 79.8 & 60.2 & 89.7 & \cellcolor{blue!25}91.7 & 80.9 & 81.3 & 82.7 \\
 & {MTL}$_{+tgt+old_{5}}$ & 94.2 & 80.3 & \cellcolor{blue!25}61.5 & \cellcolor{blue!25}92.4 & 90.8 & 80.7 & 81.5 & 83.3 \\
 & MTL$_{+tgt}$ & 92.9 & 73.9 & 59.8 & 92.2 & 77.4 & 73.8 & 75.5 & 78.3 \\
 \midrule
NER & MTL$_{-1}$  & 94.0 & 82.1 & \cellcolor{blue!25}62.2 & \cellcolor{blue!25}92.7 & - & \cellcolor{blue!25}\textbf{82.6} & 82.7 & - \\
 & {MTL}$_{+tgt+old_{10}}$ & 93.4 & 80.8 & 61.0 & 91.2 & 91.4 & 81.0 & 81.5 & 83.1 \\
 & {MTL}$_{+tgt+old_{5}}$ & 93.8 & 80.5 & \cellcolor{blue!25}62.3 & 92.1 & \cellcolor{blue!25}91.6 & 80.4 & 81.8 & 83.4 \\
 & {MTL}$_{+tgt+old+min_{10}}$ & 93.7 & 79.8 & 60.5 & 91.3 & \cellcolor{blue!25}92.0 & 81.8 & 81.4 & 83.2 \\
 & {MTL}$_{+tgt+old+min_{5}}$ & 93.6 & 80.3 & \cellcolor{blue!25}62.2 & 91.0 & 90.8 & 81.6 & 81.7 & 83.2 \\
 & MTL$_{+tgt}$ & 93.1 & 74.1 & 59.9 & 74.5 & 92.1 & 71.6 & 74.6 & 77.6 \\
 \midrule
QA & MTL$_{-1}$  & \cellcolor{blue!25}\textbf{94.1} & 81.6 & \cellcolor{blue!25}62.9 & \cellcolor{blue!25}\textbf{93.4} & \cellcolor{blue!25}92.0 & - & 84.8 & - \\
 & {MTL}$_{+tgt+old_{10}}$ & 93.5 & 80.0 & 59.7 & 91.4 & 89.9 & 81.2 & 82.9 & 82.6 \\
 & {MTL}$_{+tgt+old_{5}}$ & 94.0 & 81.1 & \cellcolor{blue!25}62.1 & 91.2 & 90.8 & 81.6 & 83.8 & 83.5 \\
 & MTL$_{+tgt}$ & 92.4 & 77.5 & 61.0 & 73.0 & 63.5 & 82.5 & 73.5 & 75.0 \\
 \bottomrule
\end{tabular}
\caption{The table reports performance score on in-language (en). Row {MTL}$_{+tgt+old_{10}}$ and{MTL}$_{+tgt+old_{5}}$ denotes adding 10\% and 5\% of existing task data combine with new task dataset respectively.{MTL}$_{+tgt+old+min_{10}}$ denotes combining the existing task dataset size minimum(10\% data , target task dataset size) i.e. to ensure the existing task dataset is less or equal to new task dataset when combined. similarly {MTL}$_{+tgt+old+min_{5}}$ denote combining the existing task dataset size as minimum(5\% data , target task dataset size). Here, column "\%$\uparrow$FLOs" reports the relative percent increase in the total computation cost for adding all 6 task with respect to the total computation cost of fine-tuning.  "Avg -1 " reports the in-language performance averaged over the task included in MTL$_{-1}$ step. "Avg" reports the cross-lingual performance averaged over all 6 task. Note, we have {MTL}$_{+tgt+old+min_{10}}$ and {MTL}$_{+tgt+old+min_{5}}$ only for NER and COPA dataset, as dataset size for NER and COPA is less. Here, text bold indicates the best value in the column and colored cell represent MTL is performing better than the Best Adapter method.}
\label{table:MTL-en-new}
\end{table*}
\endgroup

\begingroup
\setlength{\tabcolsep}{5pt} 
\renewcommand{\arraystretch}{1} 
\begin{table*}[]
\small
\centering
\begin{tabular}{l>{\columncolor[gray]{0.9}}ccccccccccccc}
\toprule
\textbf{Method} & \textbf{en} & \textbf{as} & \textbf{bn} & \textbf{gu} & \textbf{hi} & \textbf{kn} & \textbf{ml} & \textbf{mr} & \textbf{or} & \textbf{pa} & \textbf{ta} & \textbf{te} & \textbf{Avg.XL} \\
\midrule
Houlsby & \textbf{94.0} & 87.7 & 90.7 & 89.5 & 92.0 & 90.5 & 88.3 & 89.4 & 89.7 & 92.0 & 88.0 & 89.0 & 89.7 \\
Bapna & 93.3 & 87.0 & 90.1 & 89.2 & 91.8 & 89.1 & 87.0 & 88.6 & 88.5 & 90.2 & 88.1 & 88.9 & 89.0 \\
Houlsby Parallel & 93.1 & 88.1 & \textbf{91.8} & 90.5 & 93.0 & 91.0 & 88.7 & 90.3 & 90.3 & 91.7 & 88.2 & 90.0 & 90.3 \\
Bapna Parallel & 93.1 & 87.1 & 91.1 & 90.1 & 92.7 & 90.0 & 88.5 & 89.3 & 90.2 & 91.2 & 88.3 & 90.2 & 89.9 \\
Prefix Tuning & 93.8 & 85.1 & 88.9 & 88.4 & 91.7 & 88.7 & 86.2 & 89.0 & 87.7 & 90.3 & 85.7 & 88.9 & 88.2 \\
Lora & 93.4 & 83.6 & 86.3 & 85.5 & 85.3 & 85.6 & 82.3 & 86.8 & 87.1 & 86.2 & 86.1 & 88.4 & 85.7 \\
Compacter & 92.8 & 87.0 & 90.1 & 89.0 & 89.2 & 89.2 & 88.1 & 86.2 & 88.0 & 89.6 & 87.8 & 89.2 & 88.5 \\
Adapter Drop & 92.7 & 85.5 & 89.3 & 88.2 & 89.1 & 87.0 & 86.2 & 87.3 & 88.4 & 90.3 & 86.8 & 88.1 & 87.8 \\
Adapter Fusion & 93.2 & 87.4 & 91.0 & 89.6 & 91.3 & 89.9 & 87.1 & 88.7 & 89.2 & 91.4 & 87.7 & 88.7 & 89.3 \\
MADX - en & 93.6 & 88.2 & 90.4 & 89.0 & 91.4 & 90.3 & 88.4 & 89.5 & 88.9 & 91.8 & 89.1 & 89.0 & 89.6 \\
MADX - hi & 93.0 & 86.2 & 88.1 & 88.2 & 89.0 & 87.8 & 87.9 & 89.2 & 90.1 & 91.0 & \textbf{88.6} & 88.1 & 88.6 \\
\midrule
FT & 93.8 & \textbf{89.3} & 91.7 & \textbf{91.8} & \textbf{93.2} & \textbf{91.7} & \textbf{89.1} & \textbf{91.4} & 90.3 & \textbf{92.4} & 88.2 & \textbf{91.1} & \textbf{90.9} \\
MTL& 93.5 & 87.2 & 90.2 & 90.5 & 92.9 & 89.4 & 87.8 & 90.6 & \textbf{90.6} & 91.3 & 86.0 & 88.9 & 90.2 \\
\bottomrule
\end{tabular}
\caption{Results on IndicSentiment with IndicBERT. Metric: Accuracy. Column "Avg.XL" reports average cross-lingual zero-shot performance.}
\label{tab-score:sentiment}
\end{table*}
\endgroup

\begingroup
\setlength{\tabcolsep}{4pt} 
\renewcommand{\arraystretch}{1} 
\begin{table*}[]
\small
\centering
\begin{tabular}{l>{\columncolor[gray]{0.9}}ccccccccccccc}
\toprule
\textbf{Method} & \textbf{en} & \textbf{as} & \textbf{bn} & \textbf{gu} & \textbf{hi} & \textbf{kn} & \textbf{ml} & \textbf{mr} & \textbf{or} & \textbf{pa} & \textbf{ta} & \textbf{te} & \textbf{Avg.XL} \\
\midrule
Houlsby & 82.4 & 69.3 & 74.0 & 73.3 & 75.2 & 74.1 & 72.9 & 70.6 & 71.5 & 74.6 & 73.4 & 72.8 & \textbf{72.9} \\
Bapna & 81.9 & 68.7 & 73.3 & 71.1 & 73.4 & 73.3 & 73.0 & 69.3 & 71.5 & 73.9 & 72.9 & 72.8 & 72.1 \\
Houlsby Parallel & 82.5 & 69.2 & 73.0 & 72.7 & 73.8 & 73.6 & 72.6 & 70.2 & 71.7 & 74.4 & 72.6 & 72.7 & 72.4 \\
Bapna Parallel & 82.7 & 69.7 & 73.7 & 71.9 & 73.2 & 73.5 & 73.0 & 69.7 & \textbf{72.2} & 74.2 & \textbf{73.5} & 73.0 & 72.5 \\
Prefix Tuning & 82.6 & \textbf{70.9} & \textbf{74.3} & \textbf{73.7} & \textbf{75.6} & 74.0 & \textbf{73.5} & 72.1 & 72.6 & \textbf{75.2} & 73.1 & \textbf{73.4} & 73.5 \\
Lora & 80.3 & 68.1 & 71.7 & 69.8 & 72.5 & 72.2 & 70.5 & 68.5 & 69.5 & 72.7 & 70.7 & 71.3 & 70.7 \\
Compacter & 74.8 & 68.1 & 71.0 & 70.5 & 72.1 & 70.2 & 69.1 & 66.7 & 69.6 & 71.8 & 69.9 & 69.7 & 69.9 \\
Adapter Drop & 80.6 & 69.7 & 71.8 & 71.7 & 74.4 & 73.1 & 72.1 & 70.0 & 71.1 & 73.7 & 72.5 & 72.4 & 72.0 \\
Adapter Fusion & 79.9 & 68.0 & 71.6 & 70.2 & 72.8 & 71.9 & 71.8 & 68.2 & 70.0 & 73.0 & 70.9 & 69.9 & 70.8 \\
MADX -en & 82.1 & 69.9 & 73.3 & 72.9 & 73.8 & 73.0 & 72.4 & 69.5 & 70.7 & 74.2 & 73.2 & 73.0 & 72.4 \\
MADX - hi & 79.3 & 68.3 & 72.1 & 70.1 & 72.5 & 71.5 & 70.5 & 68.7 & 70.3 & 73.2 & 70.9 & 71.1 & 70.8 \\
\midrule
FT & \textbf{83.0} & 69.4 & 73.1 & 73.5 & 75.1 & \textbf{74.4} & 72.6 & \textbf{71.0} & 71.4 & 75.0 & 72.8 & 73.0 & \textbf{72.9} \\
MTL & 80.9 & 67.4 & 71.9 & 71.5 & 72.9 & 71.7 & 69.9 & 68.9 & 69.8 & 72.9 & 70.0 & 70.6 & 70.7
 \\
\bottomrule
\end{tabular}
\caption{Results on IndicXNLI task with IndicBERT. Metric: Accuracy. Column "Avg.XL" reports average cross-lingual zero-shot performance.}
\label{tab-score:xnli}
\end{table*}
\endgroup

\begingroup
\setlength{\tabcolsep}{5pt} 
\renewcommand{\arraystretch}{1} 
\begin{table*}[]
\small
\centering
\begin{tabular}{l>{\columncolor[gray]{0.9}}ccccccccccccc}
\toprule
\textbf{Method} & \textbf{en} & \textbf{as} & \textbf{bn} & \textbf{gu} & \textbf{hi} & \textbf{kn} & \textbf{ml} & \textbf{mr} & \textbf{or} & \textbf{pa} & \textbf{ta} & \textbf{te} & \textbf{Avg.XL} \\
\midrule
Houlsby & 61.5 & 63.0 & 66.4 & 64.7 & 66.4 & 63.8 & 62.2 & 63.3 & 59.2 & 63.2 & 66.2 & 66.6 & 64.1 \\
Bapna & 59.9 & 61.4 & 66.4 & 60.5 & 58.6 & 60.2 & 59.2 & 62.1 & 59.6 & 59.6 & 61.4 & 60.4 & 60.9 \\
Houlsby Parallel & 61.4 & 61.2 & 65.6 & 63.2 & 61.7 & 62.2 & 62.6 & 65.5 & 60.8 & 63.6 & 68.0 & \textbf{66.8} & 63.7 \\
Bapna Parallel & 60.5 & 60.4 & 63.0 & 61.6 & 59.0 & 60.4 & 61.4 & 64.4 & 59.8 & 60.2 & 64.2 & 61.4 & 61.4 \\
Prefix Tuning & 61.1 & 62.2 & 65.6 & \textbf{67.4} & \textbf{66.8} & 66.6 & 61.8 & 61.9 & 65.2 & 64.2 & \textbf{69.6} & 67.2 & 65.3 \\
Lora & 57.4 & 60.0 & 64.2 & 58.7 & 62.1 & 64.6 & 60.0 & 61.5 & 58.0 & 58.4 & 59.2 & 60.8 & 60.7 \\
Compacter & 50.8 & 59.8 & 66.6 & 62.9 & 63.5 & 64.0 & 63.0 & 63.3 & 58.2 & 62.4 & 66.0 & 66.0 & 63.2 \\
Adapter Drop & 52.3 & 59.6 & 64.0 & 61.2 & 60.4 & 64.6 & 62.4 & 61.7 & 57.2 & 61.6 & 64.0 & 63.0 & 61.8 \\
Adapter Fusion & 59.9 & 57.6 & 65.2 & 58.5 & 58.4 & 58.8 & 57.8 & 61.0 & 58.8 & 59.4 & 58.6 & 58.4 & 59.3 \\
MADX -en & 56.9 & 60.8 & 65.8 & 61.8 & 63.3 & 62.8 & 57.8 & 64.8 & 60.0 & 62.8 & 63.6 & 64.8 & 62.6 \\
MADX - hi & 58.4 & 62.2 & \textbf{67.2} & 62.3 & 63.5 & 63.0 & 59.4 & 63.0 & 60.2 & 64.0 & 66.0 & 63.4 & 63.1 \\
\midrule
FT & \textbf{62.3} & 61.2 & 65.2 & 60.5 & 59.5 & 61.8 & 62.0 & 60.8 & 61.0 & 63.4 & 68.0 & 63.8 & 62.5 \\
MTL & 61.4 & \textbf{64.6} & 66.6 & 62.5 & 64.4 & \textbf{66.6} & \textbf{64.4} & \textbf{67.3} & \textbf{65.8} & \textbf{64.6} & 65.4 & 66.6 & \textbf{65.3} \\
\bottomrule
\end{tabular}
\caption{Results on IndicCOPA with IndicBERT. Metric: Accuracy. Column "Avg.XL" reports average cross-lingual zero-shot performance.}
\label{tab-score:copa}
\end{table*}
\endgroup

\begingroup
\setlength{\tabcolsep}{5pt} 
\renewcommand{\arraystretch}{1} 
\begin{table*}[]
\small
\addtolength{\tabcolsep}{1pt}
\centering
\begin{tabular}{l>{\columncolor[gray]{0.9}}cccccccccccc}
\toprule
\textbf{Method} & \textbf{en} & \textbf{as} & \textbf{bn} & \textbf{gu} & \textbf{hi} & \textbf{kn} & \textbf{ml} & \textbf{mr} & \textbf{or} & \textbf{pa} & \textbf{te} & \textbf{Avg.XL} \\
\midrule
Houlsby & 92.3 & 57.8 & 50.8 & 75.7 & 51.2 & 59.7 & 57.4 & 54.6 & 57.6 & 53.8 & 55.5 & 57.4 \\
Bapna & 91.4 & 56.5 & 49.6 & 72.6 & 49.9 & 57.2 & 56.0 & 53.0 & 55.8 & 54.0 & 54.6 & 55.9 \\
Houlsby Parallel & 90.6 & 56.6 & 49.8 & 71.2 & 50.3 & 57.2 & 55.8 & 53.1 & 55.9 & 53.9 & 54.2 & 55.8 \\
Bapna Parallel & 91.3 & 56.7 & 50.0 & 72.8 & 50.7 & 57.6 & 56.4 & 53.0 & 56.6 & 53.8 & 55.1 & 56.3 \\
Prefix Tuning & 92.2 & 55.3 & 49.1 & 73.8 & 49.7 & 55.5 & 54.8 & 53.6 & 55.2 & 57.1 & 53.7 & 55.8 \\
Lora & 90.2 & 54.8 & 50.0 & 70.0 & 50.0 & 55.8 & 54.6 & 51.8 & 53.8 & 54.9 & 53.9 & 55.0 \\
Compacter & 72.7 & 49.6 & 47.0 & 63.9 & 48.3 & 45.1 & 46.3 & 48.9 & 47.1 & 59.8 & 52.5 & 50.8 \\
Adapter Drop & 75.0 & 50.8 & 49.5 & 68.1 & 50.2 & 47.7 & 49.6 & 50.2 & 49.6 & 58.6 & 54.6 & 52.9 \\
Adapter Fusion & 92.2 & 57.1 & 49.8 & 73.5 & 50.4 & 57.0 & 56.6 & 52.9 & 56.6 & 54.2 & 55.0 & 56.3 \\
MADX -en & 91.0 & 56.5 & 49.9 & 72.5 & 50.4 & 56.5 & 55.2 & 53.2 & 55.2 & 54.9 & 54.9 & 55.9 \\
MADX - hi & 90.6 & 57.1 & 49.6 & 73.4 & 50.3 & 58.4 & 55.7 & 53.5 & 56.7 & 54.9 & 54.9 & 56.5 \\
\midrule
FT & \textbf{93.0} & 56.8 & 50.9 & 76.5 & 51.1 & 57.8 & 56.5 & 55.0 & 56.7 & 56.2 & 55.0 & 57.3 \\
MTL & 91.5 & \textbf{70.7} & \textbf{88.3} & \textbf{81.3} & \textbf{81.7} & \textbf{74.7} & \textbf{73.6} & \textbf{75.9} & \textbf{66.2} & \textbf{58.7} & \textbf{71.6} & \textbf{74.3} \\
\bottomrule
\end{tabular}
\caption{Results on IndicXParaphrase with IndicBERT. Metric: Accuracy.Column "Avg.XL" reports average cross-lingual zero-shot performance.}
\label{tab-score:paraphrase}
\end{table*}
\endgroup

\begingroup
\setlength{\tabcolsep}{5pt} 
\renewcommand{\arraystretch}{1} 
\begin{table*}[]
\small
\centering
\begin{tabular}{l>{\columncolor[gray]{0.9}}ccccccccccccc}
\toprule
\textbf{Method} & \textbf{en} & \textbf{as} & \textbf{bn} & \textbf{gu} & \textbf{hi} & \textbf{kn} & \textbf{ml} & \textbf{mr} & \textbf{or} & \textbf{pa} & \textbf{ta} & \textbf{te} & \textbf{Avg.XL} \\
\midrule
Houlsby & 91.5 & 41.7 & 69.2 & 77.5 & \textbf{78.3} & 71.8 & 77.6 & 76.5 & 16.4 & 63.9 & 68.8 & 79.1 & 65.5 \\
Bapna & 91.0 & 37.5 & 70.0 & 78.3 & 76.2 & 70.9 & 78.3 & 77.9 & 16.1 & 65.1 & 67.9 & 79.3 & 65.2 \\
Houlsby Parallel & 92.2 & \textbf{46.2} & 72.0 & 77.2 & 75.9 & 74.1 & \textbf{79.7} & 77.9 & 17.3 & 63.1 & 69.1 & \textbf{81.2} & \textbf{66.7} \\
Bapna Parallel & 91.1 & 34.6 & 70.5 & 76.9 & 75.7 & 72.3 & 78.2 & 74.8 & 16.9 & 62.6 & 69.1 & 79.6 & 64.7 \\
Prefix Tuning & 91.5 & 42.6 & \textbf{72.1} & 77.7 & 76.2 & 75.0 & \textbf{79.7} & 78.1 & \textbf{17.3} & 68.6 & 69.1 & \textbf{81.2} & 67.1 \\
Lora & 90.4 & 40.7 & 70.7 & 75.0 & 72.7 & 71.0 & 75.6 & 74.1 & 17.1 & 60.0 & 61.4 & 78.3 & 63.3 \\
Compacter & 89.2 & 38.5 & 65.8 & 73.9 & 72.6 & 67.0 & 72.5 & 71.1 & 16.6 & 59.4 & 64.1 & 73.1 & 61.3 \\
Adapter Drop & 90.4 & 30.2 & 71.2 & 76.3 & 75.4 & 71.4 & 77.4 & 78.3 & 16.8 & 66.7 & 65.4 & 79.1 & 64.4 \\
Adapter Fusion & 92.0 & 42.6 & 71.8 & \textbf{79.2} & 75.1 & 76.2 & 79.4 & 78.5 & 16.5 & 66.3 & 69.8 & 80.7 & 66.9 \\
MADX -en & 91.5 & 43.6 & 69.1 & 78.9 & 75.3 & 73.9 & 78.8 & 76.6 & 16.2 & 63.8 & 68.7 & 81.1 & 66.0 \\
MADX - hi & 91.1 & 34.6 & 69.9 & 76.6 & 75.6 & 70.8 & 76.8 & 74.2 & 17.0 & 63.5 & 66.7 & 79.3 & 64.1 \\
\midrule
FT & \textbf{92.8} & 38.7 & 71.6 & 77.4 & 77.8 & \textbf{75.3} & 79.3 & \textbf{78.7} & 17.1 & 65.6 & \textbf{70.7} & 81.6 & \textbf{66.7} \\
MTL & 91.0 & 34.0 & 69.8 & 78.3 & 76.0 & 74.2 & 77.9 & 78.4 & 16.1 & \textbf{66.9} & 66.7 & 79.7 & 65.3 \\
\bottomrule
\end{tabular}
\caption{Results on IndicNER task with IndicBERT. Metric: F1 Score. Column "Avg.XL" reports average cross-lingual zero-shot performance.}
\label{tab-score:ner}
\end{table*}
\endgroup

\begingroup
\setlength{\tabcolsep}{5pt} 
\renewcommand{\arraystretch}{1} 
\begin{table*}[]
\small
\centering
\begin{tabular}{l>{\columncolor[gray]{0.9}}ccccccccccccc}
\toprule
\textbf{Method} & \textbf{en} & \textbf{as} & \textbf{bn} & \textbf{gu} & \textbf{hi} & \textbf{kn} & \textbf{ml} & \textbf{mr} & \textbf{or} & \textbf{pa} & \textbf{ta} & \textbf{te} & \textbf{Avg.XL} \\
\midrule
Houlsby & 81.7 & \textbf{44.7} & \textbf{52.9} & \textbf{45.2} & 54.9 & 46.7 & \textbf{46.2} & \textbf{48.9} & \textbf{51.8} & 52.4 & \textbf{44.9} & \textbf{60.9} & \textbf{50.0} \\
Bapna & 80.9 & 44.1 & 51.4 & 44.0 & 55.6 & 46.4 & 42.5 & 45.9 & 49.8 & 52.4 & 43.0 & 60.0 & 48.6 \\
Houlsby Parallel & 82.0 & 43.9 & 52.6 & 44.4 & 55.2 & \textbf{47.5} & 43.9 & 46.2 & 50.7 & \textbf{53.2} & 43.6 & 60.3 & 49.2 \\
Bapna Parallel & 81.4 & 44.2 & 52.0 & 43.6 & 55.6 & 47.2 & 43.6 & 45.4 & 50.8 & 52.8 & 43.4 & 59.7 & 48.9 \\
Prefix Tuning & 81.0 & 43.0 & 50.9 & 43.9 & 52.7 & 46.8 & 43.2 & 46.5 & 51.1 & 50.8 & 43.5 & 59.9 & 48.4 \\
Lora & 79.5 & 41.9 & 50.6 & 43.9 & 52.9 & 44.3 & 43.0 & 44.3 & 48.8 & 51.4 & 43.1 & 57.8 & 47.4 \\
Compacter & 73.0 & 40.8 & 48.5 & 42.3 & 50.9 & 43.9 & 41.6 & 45.1 & 46.8 & 49.6 & 42.0 & 59.2 & 46.4 \\
Adapter Drop & 70.7 & 38.3 & 46.8 & 40.9 & 50.1 & 42.3 & 40.3 & 43.0 & 46.3 & 47.4 & 38.1 & 55.2 & 44.4 \\
Adapter Fusion & 81.9 & 44.4 & 51.9 & 43.9 & \textbf{55.8} & 46.0 & 42.8 & 45.5 & 50.1 & 52.1 & 43.6 & 59.5 & 48.7 \\
MADX-en & 81.1 & 41.4 & 50.6 & 43.3 & 53.8 & 45.3 & 42.4 & 44.8 & 49.8 & 52.1 & 42.5 & 58.1 & 47.6 \\
MADX-hi & 79.4 & 41.1 & 50.2 & 43.7 & 54.9 & 44.9 & 41.9 & 44.3 & 49.0 & 51.4 & 41.5 & 58.6 & 47.4 \\
\midrule
FT & \textbf{82.1} & 44.4 & 52.8 & 44.9 & 54.6 & 46.9 & 44.6 & 46.5 & 51.3 & 52.0 & 43.9 & 60.3 & 49.3 \\
MTL & \textbf{82.1} & 39.8 & 49.1 & 42.6 & 48.9 & 42.9 & 42.2 & 43.6 & 48.1 & 47.3 & 39.7 & 56.2 & 45.5  \\
\bottomrule
\end{tabular}
\caption{Results on IndicQA task with IndicBERT. Metric: F1 score. Column "Avg.XL" reports average cross-lingual zero-shot performance.} 
\label{tab-score:qa}
\end{table*}
\endgroup

\begin{table*}[]
\small
\centering
\begin{tabular}{l>{\columncolor[gray]{0.9}}cccccccccccc}
\toprule
\textbf{Method}& \textbf{en} & \textbf{as} & \textbf{bn} & \textbf{gu} & \textbf{hi} & \textbf{kn} & \textbf{ml} & \textbf{mr} & \textbf{or} & \textbf{pa} & \textbf{ta} & \textbf{te} \\
\midrule
Houlsby & 83.9 & 60.7 & 67.4 & \textbf{71.0} & \textbf{69.7} & 67.8 & \textbf{67.4} & \textbf{67.2} & 57.7 & 66.7 & 68.3 & 70.7 \\
Pfeiffer & 83.1 & 59.2 & 66.8 & 69.3 & 67.6 & 66.2 & 66.0 & 66.1 & 56.9 & 65.9 & 66.7 & 69.3 \\
Houlsby Parallel & 83.6 & \textbf{60.8} & 67.4 & 69.9 & 68.3 & 67.6 & 67.2 & \textbf{67.2} & 57.8 & 66.6 & 68.3 & \textbf{70.9} \\
Pfeiffer Parallel & 83.4 & 58.8 & 66.7 & 69.5 & 67.8 & 66.8 & 66.8 & 66.1 & 57.8 & 65.8 & 67.7 & 69.8 \\
Prefix Tuning & 83.7 & 59.9 & 66.8 & 70.8 & 68.8 & 67.8 & 66.5 & 66.9 & \textbf{58.2} & \textbf{67.7} & 68.2 & 70.7 \\
Lora & 81.8 & 58.2 & 65.6 & 67.1 & 65.9 & 65.6 & 64.3 & 64.5 & 55.7 & 63.9 & 64.1 & 68.4 \\
Compacter & 75.5 & 57.3 & 64.8 & 67.1 & 66.1 & 63.2 & 63.4 & 63.5 & 54.4 & 65.4 & 66.0 & 68.3 \\
Adapter Drop & 77.0 & 55.7 & 65.4 & 67.7 & 66.6 & 64.3 & 64.7 & 65.1 & 54.9 & 66.4 & 65.3 & 68.8 \\
Adapter Fusion & 83.2 & 59.5 & 66.9 & 69.1 & 67.3 & 66.6 & 65.9 & 65.8 & 56.9 & 66.1 & 66.1 & 68.7 \\
MADX - en & 82.7 & 60.1 & 66.5 & 69.7 & 68.0 & 67.0 & 65.8 & 66.4 & 56.8 & 66.6 & 67.4 & 70.1 \\
MADX - hi & 82.0 & 58.3 & 66.2 & 69.0 & 67.7 & 66.1 & 65.4 & 65.5 & 57.2 & 66.3 & 66.7 & 69.2\\
FT & \textbf{84.5} & 60.0 & \textbf{67.6} & 70.8 & 68.6 & \textbf{68.0} & \textbf{67.4} & \textbf{67.2} & 58.0 & 67.4 & \textbf{68.7} & 70.8\\
\bottomrule
\end{tabular}
\caption{This table compares the performance of various adapters and FT with results averaged across all tasks. }
\label{tab-score:adapter-vs-language}
\end{table*}

\begin{table*}[]
\small
\centering
\begin{tabular}{lcccc} 
\toprule
\textbf{Method} & \textbf{FLOP} & \textbf{\% $\uparrow$ FLOS} & \textbf{Epoch} & \textbf{\% $\uparrow$ Epoch} \\
\midrule
Houlsby & 1.8E+17 & 249.8 & 17 & 240 \\
Bapna & 1.5E+17 & 185.2 & 14 & 180 \\
Houlsby Parallel & 1.1E+17 & 105.3 & 10 & 100 \\
Bapna Parallel & 8.6E+16 & 62.9 & 8 & 60 \\
Prefix Tuning & 1.5E+17 & 190.9 & 13 & 160 \\
Lora & 1.7E+17 & 223.2 & 16 & 220 \\
Compacter & 2.4E+17 & 363.9 & 23 & 360 \\
Adapter Drop & 1.2E+17 & 124.3 & 11 & 120 \\
FT & 5.3E+16 & 0.0 & 5 & 0 \\
Avg Adapter & 1.5E+17 & 188.2 & 14 & 180 \\
\bottomrule
\end{tabular}
\caption{This table report the total computation cost on Sentiment task for FT and various adapters using IndicBERT. Here \% $\uparrow$ FLOS refers to the percent increase of FLOs relative to FLOs of FT, similarly \% $\uparrow$ Epoch reports percent increase of epoch relative to epoch of FT}
\label{tab-compute:sentiment}
\end{table*}

\begin{table*}[]
\small
\centering
\begin{tabular}{lcccc}
\toprule
\textbf{Method} & \textbf{FLOP} & \textbf{\% $\uparrow$ FLOS} & \textbf{Epoch} & \textbf{\% $\uparrow$ Epoch} \\
\midrule
Houlsby & 4.0E+17 & 208.5 & 15 & 200 \\
Bapna & 4.5E+17 & 246.5 & 17 & 240 \\
Houlsby Parallel & 4.0E+17 & 208.5 & 15 & 200 \\
Bapna Parallel & 3.9E+17 & 205.4 & 15 & 200 \\
Prefix Tuning & 4.4E+17 & 237.2 & 15 & 200 \\
Lora & 3.9E+17 & 203.1 & 15 & 200 \\
Compacter & 2.9E+17 & 121.7 & 11 & 120 \\
Adapter Drop & 4.2E+17 & 225.6 & 16 & 220 \\
FT & 1.3E+17 & 0.0 & 5 & 0 \\
Average Adapter & 4.0E+17 & 207.1 & 14.88 & 197.5 \\
\bottomrule
\end{tabular}
\caption{This table report the total computation cost on XNLI task for FT and various adapters using IndicBERT.. here \% $\uparrow$ FLOS refers to the percent increase of FLOs relative to FLOs of FT, similarly \% $\uparrow$ Epoch reports percent increase of epoch relative to epoch of FT}
\label{tab-compute:XNLI}
\end{table*}

\begin{table*}[]
\small
\centering
\begin{tabular}{lcccc}
\toprule
\textbf{Method} & \textbf{FLOP} & \textbf{\% $\uparrow$ FLOS} & \textbf{Epoch} & \textbf{\% $\uparrow$ Epoch} \\
\midrule
Houlsby & 3.8E+17 & 376.6 & 28 & 366.7 \\
Bapna & 3.3E+17 & 321.0 & 25 & 316.7 \\
Houlsby Parallel & 3.0E+17 & 274.2 & 22 & 266.7 \\
Bapna Parallel & 2.3E+17 & 185.7 & 17 & 183.3 \\
Prefix Tuning & 2.2E+17 & 179.4 & 15 & 150.0 \\
Lora & 2.1E+17 & 168.0 & 16 & 166.7 \\
Compacter & 5.9E+17 & 650.9 & 45 & 650.0 \\
Adapter Drop & 1.9E+17 & 136.4 & 14 & 133.3 \\
FT & 7.9E+16 & 0.0 & 6 & 0.0 \\
Average Adapter & 3.1E+17 & 286.5 & 22.75 & 279.2 \\
\bottomrule
\end{tabular}
\caption{This table report the total computation cost on COPA task for FT and various adapters using IndicBERT.. here \% $\uparrow$ FLOS refers to the percent increase of FLOs relative to FLOs of FT, similarly \% $\uparrow$ Epoch reports percent increase of epoch relative to epoch of FT}
\label{tab-compute:COPA}
\end{table*}

\begin{table*}[]
\small
\centering
\begin{tabular}{lcccc}
\toprule
\textbf{Method} & \textbf{FLOP} & \textbf{\% $\uparrow$ FLOS} & \textbf{Epoch} & \textbf{\% $\uparrow$ Epoch} \\
\midrule
Houlsby & 1.5E+17 & 88.5 & 22 & 83.3 \\
Bapna & 1.1E+17 & 43.6 & 17 & 41.7 \\
Houlsby Parallel & 7.4E+16 & -5.8 & 11 & -8.3 \\
Bapna Parallel & 1.2E+17 & 52.6 & 18 & 50.0 \\
Prefix Tuning & 1.5E+17 & 96.2 & 21 & 75.0 \\
Lora & 1.5E+17 & 93.6 & 23 & 91.7 \\
Compacter & 9.8E+16 & 25.9 & 15 & 25.0 \\
Adapter Drop & 4.6E+16 & -40.6 & 7 & -41.7 \\
FT & 7.8E+16 & 0.0 & 12 & 0.0 \\
Total Adapter & 1.1E+17 & 44.2 & 16.75 & 39.6 \\
\bottomrule
\end{tabular}
\caption{This table report the total computation cost on Paraphrase task for FT and various adapters using IndicBERT. here \% $\uparrow$ FLOS refers to the percent increase of FLOs relative to FLOs of FT, similarly \% $\uparrow$ Epoch reports percent increase of epoch relative to epoch of FT}
\label{tab-compute:paraphrase}
\end{table*}

\begin{table*}[]
\small
\centering
\begin{tabular}{lcccc}
\toprule
\textbf{Method} & \textbf{FLOP} & \textbf{\% $\uparrow$ FLOS} & \textbf{Epoch} & \textbf{\% $\uparrow$ Epoch} \\
\midrule
Houlsby & 4.2E+15 & 19.8 & 14 & 16.7 \\
Bapna & 6.2E+15 & 77.7 & 21 & 75.0 \\
Houlsby Parallel & 6.6E+15 & 88.0 & 22 & 83.3 \\
Bapna Parallel & 4.4E+15 & 26.9 & 15 & 25.0 \\
Prefix Tuning & 6.2E+15 & 77.4 & 19 & 58.3 \\
Lora & 8.5E+15 & 143.6 & 29 & 141.7 \\
Compacter & 1.2E+16 & 252.4 & 42 & 250.0 \\
Adapter Drop & 4.2E+15 & 19.8 & 14 & 16.7 \\
FT & 3.5E+15 & 0.0 & 12 & 0.0 \\
Average Adapter & 6.6E+15 & 88.2 & 22 & 83.3 \\
\bottomrule
\end{tabular}
\caption{This table report the total computation cost on NER task for FT and various adapters using IndicBERT. here \% $\uparrow$ FLOS refers to the percent increase of FLOs relative to FLOs of FT, similarly \% $\uparrow$ Epoch reports percent increase of epoch relative to epoch of FT}
\label{tab-compute:NER}
\end{table*}

\begin{table*}[]
\small
\centering
\begin{tabular}{lcccc}
\toprule
\textbf{Method} & \textbf{FLOP} & \textbf{\% $\uparrow$ FLOS} & \textbf{Epoch} & \textbf{\% $\uparrow$ Epoch} \\
\midrule
Houlsby & 7.3E+17 & 599.0 & 41 & 583.3 \\
Bapna & 5.8E+17 & 456.7 & 33 & 450.0 \\
Houlsby Parallel & 3.9E+17 & 275.0 & 22 & 266.7 \\
Bapna Parallel & 5.1E+17 & 389.4 & 29 & 383.3 \\
Prefix Tuning & 3.1E+17 & 198.1 & 16 & 166.7 \\
Lora & 5.2E+17 & 402.9 & 30 & 400.0 \\
Compacter & 8.7E+17 & 735.6 & 50 & 733.3 \\
Adapter Drop & 1.1E+17 & 1.9 & 6 & 0.0 \\
FT & 1.0E+17 & - & 6 & - \\
Avg Adapter & 5.0E+17 & 382.3 & 28.4 & 372.9 \\
\bottomrule
\end{tabular}
\caption{This table report the total computation cost on QA task for FT and various adapters using IndicBERT. here \% $\uparrow$ FLOS refers to the percent increase of FLOs relative to FLOs of FT, similarly \% $\uparrow$ Epoch reports percent increase of epoch relative to epoch of FT}
\label{tab-compute:QA}
\end{table*}

\begin{table*}[]
\small
\centering
\begin{tabular}{lcc}
\toprule
\textbf{Method} & \textbf{XLMR-Base} & \textbf{XLMR-Large} \\
\toprule
Houlsby & 484.3 & 200.5 \\
Bapna & 547.1 & 139.9 \\
Houlsby Parallel & 197.0 & 143.3 \\
Bapna Parallel & 409.1 & 168.6 \\
Prefixtuning & 256.5 & 287.3 \\
Lora & 734.7 & 270.0 \\
compacter & 805.3 & 490.1 \\
Adapter drop & 345.1 & 214.1 \\
\bottomrule
\end{tabular}
\caption{This table reports the \textbf{percentage} increase in total FLOs with respect to FT for both XLMR-Base and XLMR-Large model}
\label{compute table: main-compute-xlmr}
\end{table*}

\begin{table*}[]
\small
\centering
\begin{tabular}{l|cccc|cccc}
\toprule
 & \multicolumn{4}{c}{\textbf{XLMR-Base}} & \multicolumn{4}{c}{\textbf{XLMR-Large}} \\
\midrule
\textbf{Method} & \textbf{WikiANN} & \textbf{XNLI} & \textbf{XQuAD} & \textbf{Total} & \textbf{WikiANN} & \textbf{XNLI} & \textbf{XQuAD} & \textbf{Total} \\
\midrule
Houlsby & 506.4 & 483.1 & 484.0 & 484.3 & 316.4 & 75.6 & 299.6 & 200.5 \\
Bapna & 439.4 & 464.6 & 582.0 & 547.1 & 172.3 & 73.7 & 194.3 & 139.9 \\
Houlsby Parallel & 270.3 & 483.1 & 86.0 & 197.0 & 103.8 & 126.8 & 160.0 & 143.3 \\
Pfeiffer Parallel & 643.4 & 409.2 & 401.0 & 409.1 & 71.1 & 167.5 & 175.9 & 168.6 \\
Prefixtuning & 157.4 & 237.7 & 267.0 & 256.5 & 200.6 & 226.3 & 344.9 & 287.3 \\
Lora & 474.3 & 404.0 & 869.0 & 734.7 & 332.1 & 57.9 & 446.9 & 270.0 \\
Compacter & 905.8 & 818.2 & 797.0 & 805.3 & 528.9 & 336.8 & 618.4 & 490.1 \\
Adapter drop & 281.9 & 409.2 & 323.0 & 345.1 & 617.0 & 153.1 & 240.0 & 214.1 \\
\bottomrule
\end{tabular}
\caption{This table reports the \textbf{percentage} increase in computational cost with respect to FT for XLM-R model for task NER, XNLI and QA. "Total" reports the percentage increase of total FLOs for the method relative to total FT FLOs}
\label{compute table: Overall XLMR}
\end{table*}

\begin{table*}[]
\addtolength{\tabcolsep}{-1pt}
\small
\centering
\begin{tabular}{l|cccc|cccc}
\toprule
\textbf{ EN} & \multicolumn{4}{c}{\textbf{XLMR-Base}} & \multicolumn{4}{|c}{\textbf{XLMR-Large}} \\
\midrule
\textbf{Method} & \textbf{NER} & \textbf{XNLI} & \textbf{QA} & \textbf{Average} & \textbf{NER} & \textbf{XNLI} & \textbf{QA} & \textbf{Average} \\
\midrule
Houlsby & 81.0 & 82.7 & \textbf{84.1} & 82.6 & \textbf{83.5} & 85.9 & 88.2 & 85.9 \\
Bapna & 79.9 & 81.3 & 83.2 & 81.5 & 83.1 & 86.4 & 87.4 & 85.7 \\
Houlsby parallel & 80.5 & 83.5 & 83.7 & 82.6 & 83.0 & 87.9 & 88.0 & 86.3 \\
Bapna parallel & 80.8 & 80.9 & 82.8 & 81.5 & 82.5 & 88.0 & 87.7 & 86.1 \\
Prefixtuning & 79.0 & 79.5 & 81.7 & 80.1 & 83.2 & \textbf{88.2} & \textbf{88.3} & \textbf{86.5} \\
Lora & 78.6 & 79.7 & 81.7 & 80.0 & 81.7 & 85.6 & 86.9 & 84.7 \\
compacter & 72.3 & 76.4 & 76.6 & 75.1 & 76.1 & 85.6 & 85.0 & 82.2 \\
Adapter drop & 81.1 & 80.3 & 82.7 & 81.4 & 82.6 & 88.0 & 88.0 & 86.2 \\
FT & \textbf{82.3} & \textbf{83.1} & 83.3 & \textbf{82.9} & 82.8 & 87.3 & 88.0 & 86.0 \\
\bottomrule
\end{tabular}
\caption{Overall performance on \textbf{English} for XLMR-B and XLMR-L model}
\label{Overall EN: XLMR}
\end{table*}

\begin{table*}[]
\addtolength{\tabcolsep}{-1pt}
\small
\centering
\begin{tabular}{l|cccc|cccc}
\toprule
\textbf{XL} & \multicolumn{4}{c}{\textbf{XLMR-Base}} & \multicolumn{4}{|c}{\textbf{XLMR-Large}} \\
\midrule
\textbf{Method} & \textbf{NER} & \textbf{XNLI} & \textbf{QA} & \textbf{Average} & \textbf{NER} & \textbf{XNLI} & \textbf{QA} & \textbf{Average} \\
\midrule
Houlsby & 61.0 & 72.6 & \textbf{71.5} & \textbf{68.4} & 64.6 & 76.2 & \textbf{78.6} & 73.1 \\
Bapna & 58.3 & 71.3 & 69.9 & 66.5 & 64.3 & 76.7 & 78.0 & 73.0 \\
Houlsby Parallel & 59.2 & 72.8 & 70.1 & 67.4 & \textbf{65.3} & 78.7 & 77.8 & 73.9 \\
Bapna Parallel & 57.1 & 70.3 & 69.7 & 65.7 & 63.1 & \textbf{78.8} & 77.6 & 73.2 \\
Prefixtuning & 58.5 & 69.9 & 67.7 & 65.4 & 64.7 & 78.7 & 77.6 & 73.7 \\
Lora & 58.6 & 70.5 & 68.4 & 65.8 & 62.3 & 76.9 & 77.1 & 72.1 \\
Compacter & 55.1 & 66.8 & 64.1 & 62.0 & 58.5 & 76.4 & 75.3 & 70.1 \\
Adapter drop & 60.5 & 70.2 & 71.3 & 67.3 & 64.6 & \textbf{78.8} & 78.5 & \textbf{74.0} \\
FT & \textbf{61.7} & \textbf{73.7} & 70.8 & \textbf{68.7} & 63.9 & 77.0 & 78.0 & 73.0 \\
\bottomrule
\end{tabular}
\caption{Overall \textbf{cross-lingual} performance for XLMR-B and XLMR-L model}
\label{Overall XL: XLMR}
\end{table*}

\begin{table*}[]
\small
\addtolength{\tabcolsep}{-3pt}
\centering
\begin{tabular}{l>{\columncolor[gray]{0.9}}cccccccccccccccc}
\toprule
\textbf{Method} & \textbf{en} & \textbf{ar} & \textbf{bg} & \textbf{de} & \textbf{el} & \textbf{es} & \textbf{fr} & \textbf{hi} & \textbf{ru} & \textbf{sw} & \textbf{th} & \textbf{tr} & \textbf{ur} & \textbf{vi} & \textbf{zh} & \textbf{Avg.XL} \\
\midrule
houlsby & 83.5 & 45.6 & 81.6 & 79.3 & 79.9 & 76.2 & 78.7 & 71.1 & 68.0 & 68.2 & 0.6 & 82.0 & 69.1 & 77.5 & 26.2 & 64.6 \\
Bapna & 83.1 & 41.0 & 81.4 & 78.1 & 77.2 & 77.0 & 78.7 & 73.2 & 71.5 & 68.3 & 2.0 & 79.3 & 69.1 & 76.8 & 26.2 & 64.3 \\
houlsby parallel & 83.0 & 46.4 & 83.1 & 79.2 & 79.2 & 76.1 & 79.0 & 70.0 & 71.4 & 70.4 & 0.6 & 82.0 & 75.6 & 76.6 & 24.7 & 65.3 \\
Bapna parallel & 82.5 & 48.3 & 79.3 & 77.9 & 77.9 & 72.7 & 78.3 & 66.5 & 71.5 & 68.8 & 1.4 & 80.0 & 63.3 & 75.0 & 22.2 & 63.1 \\
prefixtuning & 83.2 & 48.5 & 79.2 & 77.8 & 79.1 & 76.8 & 79.8 & 73.5 & 69.1 & 66.5 & 4.3 & 79.9 & 71.1 & 75.6 & 24.6 & 64.7 \\
lora & 81.7 & 46.0 & 80.0 & 77.9 & 76.9 & 68.7 & 77.8 & 66.9 & 67.9 & 66.6 & 2.5 & 76.4 & 65.7 & 76.8 & 21.5 & 62.3 \\
compacter & 76.1 & 38.8 & 75.5 & 75.5 & 74.8 & 73.8 & 74.8 & 62.7 & 58.7 & 60.2 & 1.2 & 75.9 & 65.5 & 69.1 & 12.2 & 58.5 \\
Adapter drop & 82.6 & 48.0 & 82.0 & 78.5 & 78.7 & 75.0 & 79.8 & 68.0 & 69.4 & 68.1 & 1.0 & 80.0 & 75.0 & 76.4 & 24.0 & 64.6 \\
FT & 82.8 & 49.3 & 81.6 & 79.1 & 76.6 & 77.7 & 81.1 & 70.6 & 70.9 & 66.9 & 0.4 & 78.3 & 60.7 & 77.7 & 23.1 & 63.9
\\
\bottomrule
\end{tabular}

\caption{Results on WikiANN task with \textbf{XLM-R Large} model, metric: F1 score}
\label{Score:XLMRL-NER}
\end{table*}

\begin{table*}[]
\small
\addtolength{\tabcolsep}{-3pt}
\centering
\begin{tabular}{l>{\columncolor[gray]{0.9}}cccccccccccccccc}
\toprule
\textbf{Method} & \textbf{en} & \textbf{ar} & \textbf{bg} & \textbf{de} & \textbf{el} & \textbf{es} & \textbf{fr} & \textbf{hi} & \textbf{ru} & \textbf{sw} & \textbf{th} & \textbf{tr} & \textbf{ur} & \textbf{vi} & \textbf{zh} & \textbf{Avg.XL} \\
\midrule
houlsby & 81.0 & 44.4 & 76.0 & 73.6 & 74.3 & 71.6 & 76.1 & 70.1 & 61.7 & 69.3 & 1.5 & 75.8 & 65.5 & 68.9 & 25.2 & 61.0 \\
Bapna & 79.9 & 41.7 & 73.2 & 72.3 & 73.0 & 73.7 & 74.9 & 62.9 & 59.2 & 67.6 & 2.0 & 72.6 & 56.2 & 62.7 & 23.6 & 58.3 \\
houlsby parallel & 80.5 & 44.7 & 75.8 & 73.3 & 73.8 & 67.7 & 74.7 & 66.4 & 62.1 & 66.7 & 1.8 & 74.0 & 57.5 & 64.8 & 26.0 & 59.2 \\
Bapna parallel & 80.8 & 42.1 & 74.3 & 72.4 & 70.6 & 70.3 & 74.3 & 62.0 & 61.1 & 61.7 & 1.0 & 71.3 & 51.3 & 62.6 & 24.4 & 57.1 \\
prefixtuning & 79.0 & 46.5 & 75.9 & 70.0 & 70.6 & 72.8 & 74.8 & 62.4 & 59.7 & 62.3 & 1.1 & 70.8 & 64.2 & 67.4 & 20.2 & 58.5 \\
lora & 78.6 & 42.8 & 74.6 & 71.2 & 70.7 & 71.7 & 74.1 & 62.4 & 57.8 & 67.0 & 2.9 & 70.6 & 63.0 & 68.0 & 23.7 & 58.6 \\
compacter & 72.3 & 42.0 & 72.9 & 70.2 & 68.3 & 61.6 & 67.6 & 59.9 & 54.4 & 62.8 & 0.7 & 69.0 & 56.8 & 63.5 & 21.6 & 55.1 \\
Adapter drop & 81.1 & 45.7 & 76.8 & 73.8 & 74.5 & 69.2 & 74.7 & 66.1 & 62.4 & 66.0 & 1.9 & 74.8 & 65.7 & 67.4 & 27.1 & 60.5 \\
FT & 82.3 & 48.5 & 77.0 & 73.3 & 74.7 & 75.3 & 75.7 & 67.7 & 63.0 & 69.2 & 3.8 & 76.6 & 64.7 & 69.8 & 24.1 & 61.7\\
\bottomrule
\end{tabular}
\caption{Results on WikiANN task with \textbf{XLM-R Base} model, metric: F1 score}
\label{Score:XLMRB-NER}
\end{table*}

\begin{table*}[]
\small
\addtolength{\tabcolsep}{-3pt}
\centering
\begin{tabular}{l>{\columncolor[gray]{0.9}}cccccccccccccccc}
\toprule
\textbf{Method} & \textbf{en} & \textbf{ar} & \textbf{bg} & \textbf{de} & \textbf{el} & \textbf{es} & \textbf{fr} & \textbf{hi} & \textbf{ru} & \textbf{sw} & \textbf{th} & \textbf{tr} & \textbf{ur} & \textbf{vi} & \textbf{zh} & \textbf{Avg.XL} \\
\midrule
houlsby & 85.9 & 75.0 & 80.0 & 80.4 & 78.8 & 81.2 & 80.0 & 73.4 & 76.8 & 69.7 & 74.1 & 76.2 & 68.3 & 76.8 & 75.4 & 76.2 \\
Bapna & 86.4 & 75.5 & 80.3 & 81.0 & 79.2 & 81.2 & 80.6 & 73.8 & 77.8 & 69.7 & 74.9 & 76.3 & 70.4 & 77.1 & 76.3 & 76.7 \\
houlsby parallel & 87.9 & 77.7 & 82.3 & 82.6 & 80.9 & 83.7 & 82.3 & 77.1 & 79.6 & 71.2 & 77.0 & 78.3 & 72.1 & 78.5 & 78.8 & 78.7 \\
Bapna parallel & 88.0 & 78.4 & 82.9 & 82.7 & 81.2 & 84.0 & 82.7 & 76.1 & 79.5 & 71.3 & 76.4 & 78.4 & 72.2 & 78.7 & 78.1 & 78.8 \\
prefixtuning & 88.2 & 78.4 & 82.3 & 81.6 & 81.6 & 83.3 & 82.7 & 76.0 & 79.6 & 71.1 & 77.3 & 78.5 & 72.6 & 78.9 & 78.3 & 78.7 \\
lora & 85.6 & 75.3 & 80.8 & 80.5 & 79.7 & 81.7 & 80.9 & 74.5 & 78.5 & 70.0 & 75.1 & 76.9 & 70.1 & 77.0 & 76.3 & 76.9 \\
compacter & 85.6 & 74.2 & 80.2 & 80.5 & 78.9 & 80.7 & 80.5 & 74.7 & 77.3 & 69.7 & 74.9 & 75.9 & 69.7 & 76.5 & 76.4 & 76.4 \\
Adapter drop & 88.0 & 77.7 & 82.8 & 82.5 & 82.1 & 84.0 & 82.6 & 76.1 & 80.1 & 72.2 & 76.7 & 78.8 & 71.3 & 79.0 & 77.6 & 78.8 \\
FT & 87.3 & 76.1 & 81.9 & 80.5 & 79.5 & 82.3 & 81.7 & 73.9 & 79.5 & 65.5 & 75.7 & 76.0 & 68.7 & 78.4 & 78.4 & 77.0\\
\bottomrule
\end{tabular}
\caption{Results on XNLI task with \textbf{XLM-R Large} model, metric: Accuracy}
\label{Score:XLMRL-XNLI}
\end{table*}

\begin{table*}[]
\small
\addtolength{\tabcolsep}{-3pt}
\centering
\begin{tabular}{l>{\columncolor[gray]{0.9}}cccccccccccccccc}
\toprule
\textbf{Method} & \textbf{en} & \textbf{ar} & \textbf{bg} & \textbf{de} & \textbf{el} & \textbf{es} & \textbf{fr} & \textbf{hi} & \textbf{ru} & \textbf{sw} & \textbf{th} & \textbf{tr} & \textbf{ur} & \textbf{vi} & \textbf{zh} & \textbf{Avg.XL} \\
\midrule
houlsby & 82.7 & 70.9 & 77.1 & 76.3 & 74.7 & 77.9 & 77.9 & 68.6 & 74.3 & 64.5 & 71.1 & 71.8 & 65.5 & 73.4 & 72.5 & 72.6 \\
Bapna & 81.3 & 68.8 & 75.3 & 74.2 & 73.4 & 76.7 & 75.8 & 67.6 & 73.5 & 64.0 & 69.7 & 71.2 & 64.0 & 72.8 & 71.2 & 71.3 \\
houlsby parallel & 83.5 & 70.3 & 77.0 & 75.9 & 74.8 & 78.0 & 77.7 & 69.6 & 74.7 & 65.1 & 71.3 & 71.7 & 65.4 & 74.8 & 73.1 & 72.8 \\
Bapna parallel & 80.9 & 68.0 & 74.7 & 73.2 & 72.1 & 76.2 & 75.0 & 67.2 & 72.4 & 63.2 & 68.1 & 70.1 & 62.3 & 71.5 & 69.8 & 70.3 \\
prefixtuning & 79.5 & 68.7 & 73.7 & 72.6 & 71.7 & 74.1 & 74.2 & 66.3 & 71.3 & 62.9 & 69.5 & 69.1 & 62.9 & 72.0 & 70.1 & 69.9 \\
lora & 79.7 & 68.4 & 75.0 & 73.6 & 72.2 & 75.3 & 74.5 & 66.6 & 72.2 & 63.6 & 68.1 & 70.9 & 63.6 & 71.4 & 71.3 & 70.5 \\
compacter & 76.4 & 64.1 & 70.7 & 70.6 & 69.4 & 72.8 & 71.8 & 61.7 & 70.0 & 60.4 & 63.4 & 67.4 & 59.0 & 68.1 & 66.5 & 66.8 \\
Adapter drop & 80.3 & 68.1 & 74.4 & 73.6 & 71.1 & 75.9 & 75.3 & 67.2 & 72.1 & 63.3 & 68.9 & 69.9 & 62.1 & 71.7 & 70.0 & 70.2 \\
FT & 83.1 & 71.3 & 78.0 & 76.6 & 75.3 & 78.6 & 76.9 & 71.3 & 75.4 & 64.0 & 73.0 & 73.0 & 67.5 & 75.6 & 74.7 & 73.7\\
\bottomrule
\end{tabular}
\caption{ Results on XNLI task with \textbf{XLM-R Base} model, metric: Accuracy}
\label{Score:XLMRB-XNLI}
\end{table*}

\textbf{\begin{table*}[]
\small
\centering
\begin{tabular}{l>{\columncolor[gray]{0.9}}ccccccccccccc}
\toprule
\textbf{Method} & \textbf{en} & \textbf{ar} & \textbf{de} & \textbf{el} & \textbf{es} & \textbf{hi} & \textbf{ro} & \textbf{ru} & \textbf{th} & \textbf{tr} & \textbf{vi} & \textbf{zh} & \textbf{Avg.XL} \\
\midrule
houlsby & 88.2 & 77.3 & 81.2 & 80.3 & 83.3 & 77.0 & 85.0 & 80.8 & 74.3 & 75.2 & 80.0 & 70.1 & 78.6 \\
Bapna & 87.4 & 75.7 & 79.9 & 80.5 & 82.6 & 75.6 & 84.1 & 80.7 & 75.6 & 73.9 & 79.8 & 69.7 & 78.0 \\
houlsby parallel & 88.0 & 75.3 & 81.4 & 80.4 & 81.9 & 76.2 & 84.2 & 79.9 & 74.2 & 74.2 & 79.2 & 68.7 & 77.8 \\
Bapna parallel & 87.7 & 75.2 & 80.4 & 80.4 & 82.0 & 75.6 & 84.1 & 79.9 & 73.7 & 73.7 & 79.4 & 69.4 & 77.6 \\
prefixtuning & 88.3 & 75.4 & 81.5 & 80.5 & 82.3 & 75.6 & 83.0 & 79.3 & 74.5 & 73.9 & 78.6 & 68.8 & 77.6 \\
lora & 86.9 & 75.8 & 80.6 & 78.5 & 81.2 & 75.0 & 82.5 & 79.1 & 75.2 & 72.7 & 77.9 & 69.4 & 77.1 \\
compacter & 85.0 & 73.7 & 77.7 & 77.6 & 79.5 & 74.7 & 80.9 & 78.3 & 70.8 & 70.5 & 76.9 & 68.2 & 75.3 \\
Adapter drop & 88.0 & 76.1 & 81.3 & 81.1 & 83.2 & 76.7 & 85.1 & 80.7 & 74.3 & 74.6 & 80.3 & 69.6 & 78.5 \\
FT & 88.0 & 76.3 & 80.7 & 80.3 & 81.8 & 76.2 & 84.2 & 79.6 & 75.0 & 74.4 & 79.8 & 69.7 & 78.0\\
\bottomrule
\end{tabular}
\caption{ Results on Squad, XQAUD task with \textbf{XLM-R Large} model, metric: F1 score}
\label{Score:XLMRL-QA}
\end{table*}}

\textbf{\begin{table*}[]
\small
\centering
\begin{tabular}{l>{\columncolor[gray]{0.9}}ccccccccccccc}
\toprule
\textbf{Method} & \textbf{en} & \textbf{ar} & \textbf{de} & \textbf{el} & \textbf{es} & \textbf{hi} & \textbf{ro} & \textbf{ru} & \textbf{th} & \textbf{tr} & \textbf{vi} & \textbf{zh} & \textbf{Avg.XL} \\
\midrule
houlsby & 84.1 & 67.0 & 75.0 & 73.3 & 76.8 & 69.8 & 79.0 & 72.6 & 68.3 & 66.6 & 73.8 & 64.6 & 71.5 \\
Bapna & 83.2 & 64.5 & 73.6 & 71.0 & 75.1 & 66.1 & 77.5 & 72.6 & 65.6 & 66.3 & 72.9 & 63.4 & 69.9 \\
houlsby parallel & 83.7 & 65.9 & 74.6 & 72.0 & 75.5 & 66.5 & 77.9 & 72.8 & 64.4 & 66.5 & 71.9 & 62.9 & 70.1 \\
Bapna parallel & 82.8 & 64.4 & 73.2 & 72.6 & 74.0 & 65.4 & 77.4 & 73.0 & 65.0 & 65.4 & 72.0 & 63.9 & 69.7 \\
prefixtuning & 81.7 & 63.1 & 71.3 & 70.0 & 72.1 & 64.4 & 75.3 & 70.2 & 62.6 & 63.5 & 69.9 & 62.1 & 67.7 \\
lora & 81.7 & 61.4 & 71.8 & 71.2 & 72.9 & 65.4 & 76.7 & 71.9 & 63.1 & 65.4 & 71.3 & 61.0 & 68.4 \\
compacter & 76.6 & 60.7 & 67.0 & 64.9 & 68.8 & 62.4 & 70.3 & 66.8 & 58.6 & 59.5 & 68.9 & 57.3 & 64.1 \\
Adapter drop & 82.7 & 66.6 & 74.3 & 73.6 & 75.3 & 70.2 & 77.0 & 74.6 & 68.2 & 66.8 & 74.5 & 62.8 & 71.3 \\
FT & 83.3 & 66.5 & 74.6 & 72.2 & 75.1 & 66.8 & 77.5 & 73.4 & 66.8 & 67.5 & 73.2 & 65.4 & 70.8\\
\bottomrule
\end{tabular}
\caption{ Results on SQUAD, XQUAD task with \textbf{XLM-R Base} model, metric: F1 score}
\label{Score:XLMRB-QA}
\end{table*}}

\textbf{\begin{table*}[]
\small
\centering
\begin{tabular}{l>{\columncolor[gray]{0.9}}ccccccccccccc}
\toprule
\textbf{Method} & \textbf{en} & \textbf{ar} & \textbf{de} & \textbf{el} & \textbf{es} & \textbf{hi} & \textbf{ro} & \textbf{ru} & \textbf{th} & \textbf{tr} & \textbf{vi} & \textbf{zh} & \textbf{Avg.XL} \\
\midrule
houlsby & 84.1 & 67.0 & 75.0 & 73.3 & 76.8 & 69.8 & 79.0 & 72.6 & 68.3 & 66.6 & 73.8 & 64.6 & 71.5 \\
Bapna & 83.2 & 64.5 & 73.6 & 71.0 & 75.1 & 66.1 & 77.5 & 72.6 & 65.6 & 66.3 & 72.9 & 63.4 & 69.9 \\
houlsby parallel & 83.7 & 65.9 & 74.6 & 72.0 & 75.5 & 66.5 & 77.9 & 72.8 & 64.4 & 66.5 & 71.9 & 62.9 & 70.1 \\
Bapna parallel & 82.8 & 64.4 & 73.2 & 72.6 & 74.0 & 65.4 & 77.4 & 73.0 & 65.0 & 65.4 & 72.0 & 63.9 & 69.7 \\
prefixtuning & 81.7 & 63.1 & 71.3 & 70.0 & 72.1 & 64.4 & 75.3 & 70.2 & 62.6 & 63.5 & 69.9 & 62.1 & 67.7 \\
lora & 81.7 & 61.4 & 71.8 & 71.2 & 72.9 & 65.4 & 76.7 & 71.9 & 63.1 & 65.4 & 71.3 & 61.0 & 68.4 \\
compacter & 76.6 & 60.7 & 67.0 & 64.9 & 68.8 & 62.4 & 70.3 & 66.8 & 58.6 & 59.5 & 68.9 & 57.3 & 64.1 \\
Adapter drop & 82.7 & 66.6 & 74.3 & 73.6 & 75.3 & 70.2 & 77.0 & 74.6 & 68.2 & 66.8 & 74.5 & 62.8 & 71.3 \\
FT & 83.3 & 66.5 & 74.6 & 72.2 & 75.1 & 66.8 & 77.5 & 73.4 & 66.8 & 67.5 & 73.2 & 65.4 & 70.8\\
\bottomrule
\end{tabular}
\caption{ Results on SQUAD, XQUAD task with \textbf{XLM-R Base} model, metric: F1 score}
\label{Score:XLMRB-QA}
\end{table*}}

\subsection{MTL maintainability}
MTL is maintainable as discussed in sec~\ref{sec:maintainability}, as the MTL model can be extended to new tasks by continually learning with the new task's data along with 10\% of the existing tasks' data. We analyze the impact of performance and computational cost by changing the percentage of an existing task for continual learning of new task as presented in Table~\ref{table:MTL_XL_new} and \ref{table:MTL-en-new}. We tested two additional setups: (a) using 5\% data from previously seen tasks (\textbf{MTL$_{-1}$}) instead of 10\%, as reported in the "{MTL}$_{+tgt+old_{5}}$" row and  (b) using the minimum of either 10\% of the existing task dataset or the new task dataset, reported in the row "{MTL}$_{+tgt+old+min_{10}}$", and similarly, using the minimum of either 5\% of the existing task dataset or the new task dataset, reported in the row "{MTL}$_{+tgt+old+min_{5}}$". Our findings show that cross-lingual performance is better when using a higher percentage of the existing task dataset, while in-language performance is better when using a lower percentage of the existing task dataset. In terms of computational efficiency, using 5\% of the existing dataset requires fewer FLOs compared to using 10\%.

\end{document}